\pdfoutput=1

\documentclass[11pt]{article}

\usepackage[final]{acl}

\usepackage{times}
\usepackage{latexsym}

\usepackage[T1]{fontenc}

\usepackage[utf8]{inputenc}

\usepackage{microtype}

\usepackage{inconsolata}

\usepackage{graphicx}
\usepackage{amsmath}
\usepackage{amsthm}
\newtheorem*{definition*}{Definition}
\usepackage{booktabs}
\usepackage{multirow}
\usepackage{comment}
\usepackage[normalem]{ulem}

\usepackage{titlesec}

\title{AIRepr: An Analyst-Inspector Framework for Evaluating Reproducibility of LLMs in Data Science}

\author{
  \textbf{Qiuhai Zeng\textsuperscript{1}\thanks{These authors contributed equally to this work.}} \quad
  \textbf{Claire Jin\textsuperscript{2}\footnotemark[1]} \quad
  \textbf{Xinyue Wang\textsuperscript{1}} \quad 
  \textbf{Yuhan Zheng\textsuperscript{3}} \quad
  \textbf{Qunhua Li\textsuperscript{1}\thanks{Corresponding author: qunhua.li@psu.edu.}}\\
  \\
  \textsuperscript{1}Pennsylvania State University,
  \textsuperscript{2}Carnegie Mellon University,
  \textsuperscript{3}International Monetary Fund\\
  \small{
    \texttt{\{qjz5084, xpw5228, qunhua.li\}}\href{mailto:}{\textcolor{black}{@psu.edu}}, \ 
    \href{mailto:}{\textcolor{black}{\texttt{claireji@andrew.cmu.edu}}}, \
    \href{mailto:}{\textcolor{black}{\texttt{helenzheng099@gmail.com}}}
  }
}

\begin{document}
\maketitle
\begin{abstract}

Large language models (LLMs) are increasingly used to automate data analysis through executable code generation.  Yet, data science tasks often admit multiple statistically valid solutions, e.g. different modeling strategies, making it critical to understand the reasoning behind analyses, not just their outcomes. While manual review of LLM-generated code can help ensure statistical soundness, it is labor-intensive and requires expertise. A more scalable approach is to evaluate the underlying workflows—the logical plans guiding code generation. However, it remains unclear how to assess whether an LLM-generated workflow supports reproducible implementations.

To address this, we present \textbf{AIRepr}, an {\bf A}nalyst–{\bf I}nspector framework for automatically evaluating and improving the {\bf repr}oducibility of LLM-generated data analysis workflows. Our framework is grounded in statistical principles and supports scalable, automated assessment. We introduce two novel reproducibility-enhancing prompting strategies and benchmark them against standard prompting across 15 analyst-inspector LLM pairs and 1,032 tasks from three public benchmarks. Our findings show that workflows with higher reproducibility also yield more accurate analyses, and that reproducibility-enhancing prompts substantially improve both metrics. This work provides a foundation for transparent, reliable, and efficient human-AI collaboration in data science. Our code is publicly available\footnote{\href{https://github.com/qunhualilab/LLM-DS-Reproducibility}{https://github.com/qunhualilab/LLM-DS-Reproducibility}}.

\end{abstract}

\section{Introduction}
\begin{figure*}[t]
  \includegraphics[width=\linewidth]{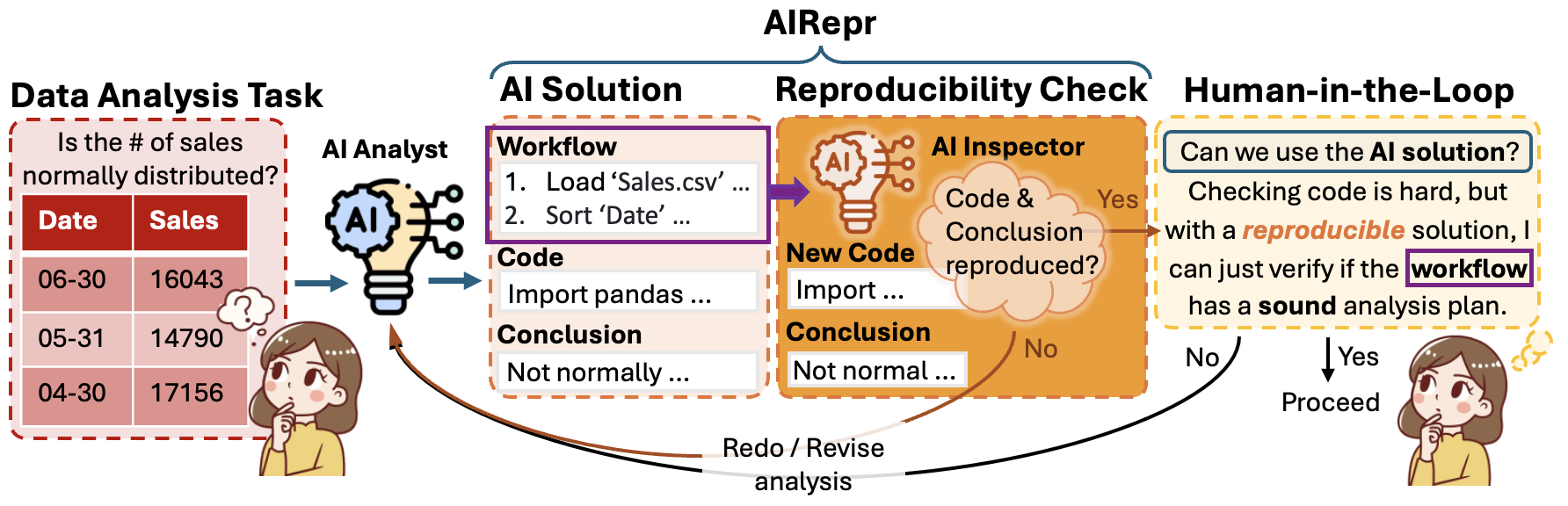} 
  \caption { 
A \textbf{human-in-the-loop pipeline} for \textbf{AI-generated data analysis}. Given a data analysis task, an AI Analyst generates a workflow (analysis steps), code, and conclusion. An independent AI Inspector takes the workflow to generate new code and a new conclusion. If the AI Analyst's conclusion is independently reproducible—meaning the workflow provides complete and sufficient details for the generated code—then human analysts can focus on evaluating the soundness of the workflow without manually verifying the code. However, if the AI solution fails the reproducibility check, the solution needs to be revised before being submitted for human review.}
  \label{fig:hci-pipeline}
\end{figure*}

Large language models (LLMs) are increasingly used to automate data analysis by generating executable code \cite{zhu2024large, gu2024analysts, nejjar2025llms, jansen2025leveraging}. With recent advances in reasoning \cite{guo2025deepseek} and tool use \cite{gao2023pal, schick2023toolformer}, LLMs are achieving impressive performance on a wide range of data science tasks, from exploratory analysis \cite{ma2023demonstration} to predictive modeling \cite{chi2024sela, jiang2025aide}, often approaching or even surpassing human-level accuracy on benchmark datasets \cite{zhu2024large}.

However, data science is not just about producing an answer. Many tasks admit multiple valid analytical paths \cite{gelman2013garden, dragicevic2019increasing}: different statistical tests, feature selection, 
 or modeling strategies can yield distinct yet  equally defensible conclusions \cite{silberzahn2018many}. Even when ground truth labels exist, rigorous analysis is necessary to avoid critical pitfalls such as missing preprocessing choices or inappropriate distributional assumptions \cite{moscovich2022cross, moreno2012unifying}. Consequently, accuracy alone is an insufficient metric of the quality of a data science solution. Understanding \textit{how} an answer is reached, i.e. the workflow of the analysis, is essential for ensuring validity, transparency, and reproducibility \cite{sandve2013ten}.

Human data scientists are trained to uphold high standards of reproducibility \cite{davidson2008provenance, NAP25104, NAP25303}, not only by writing correct code but by documenting their analytical steps and rationales (e.g. data processing steps, model specifications) in methods sections or technical reports \cite{goodman2016does, munafo2017manifesto}.  These practices reflect a broad view of reproducibility -- one that values transparency, clarity, and completeness of the information necessary for others to replicate and verify the work, just as much as correctness. Though LLMs can generate syntactically valid code and articulate underlying high-level reasoning, verifying the soundness of these outputs still demands time-intensive manual inspection by experts.

A more scalable alternative is to focus on evaluating the underlying \textit{workflow} -- the structured sequence of reasoning steps and details that compose an analysis -- over the code itself (Figure \ref{fig:hci-pipeline}). If workflows are expressed clearly and accurately, they can serve as a reproducible and inspectable artifact in their own right, enabling reviewers or collaborators to assess methodological soundness without sifting through implementation details (Example in Table \ref{WorkflowCodeExample}). However, whether LLM-derived workflows are sufficiently detailed to support this kind of inspection and independent replication remains an open question.

To bridge this gap, we propose the \textbf{analyst-inspector framework} to automatically evaluate and enhance the reproducibility of LLM-generated data analysis workflows, reducing dependence on manual code verification. In this framework, an independent inspector model attempts to reproduce the analysis based solely on the workflow produced by the analyst model. If the inspector can reproduce functionally equivalent code and arrive at the same conclusion, this indicates that the workflow contains all necessary details without relying on implicit assumptions or analyst-specific knowledge. 

Our analyst-inspector framework offers a practical and principled approach for evaluating LLM-generated data science workflows, with several key advantages. (A) \textbf{Human-aligned and intuitive}: it reflects how human reviewers typically evaluate analyses by reading methodological descriptions rather than inspecting code line-by-line. (B) \textbf{Statistically grounded}: the framework requires workflows not only to be faithful \cite{lyu2024towards}, accurately reflecting the analysis steps, but also to satisfy the classical statistical properties of sufficiency and completeness \citep{lehmann1983theory, casella2024statistical}, meaning they must include all necessary information (sufficiency) while excluding irrelevant or extraneous details (completeness).
(C) \textbf{Generalizable and lightweight}: the core mechanism of checking whether one LLM can independently reproduce another’s analysis based on its generated workflow is broadly applicable across tasks and domains, without requiring labeled training data or complex infrastructure.

Using this framework, we systematically evaluate 15 combinations of five LLMs as analysts and three LLMs as inspectors across 1,032 diverse data analysis tasks drawn from three benchmark datasets. 
We propose two novel {\bf reproducibility-enhancing prompting strategies} and benchmark them against two standard baselines to assess their impact on analysis reproducibility and accuracy.
Our study finds that increased reproducibility significantly correlates with higher accuracy in LLM-generated analyses, and that prompts explicitly emphasizing reproducibility improve analysis quality. These findings provide strong directions for advancing transparent and robust human-AI collaboration in data science. 

\section{Related work}
\subsection{LLM-Assisted Data Science}
LLMs are increasingly applied to automate data science tasks across the analytical pipeline, including data preprocessing \cite{zhang2023large}, exploratory data analysis \cite{ma2023demonstration}, visualization generation \cite{wang2025dataformulator2iterative, yang2024matplotagent}, feature engineering \cite{hollmann2023large}, and automatic machine learning \cite{guo2024ds, chi2024sela, jiang2025aide}. Early work focused on single-turn prompting to generate runnable code for straightforward questions like ``What is the correlation between X and Y?'' \cite{wei2022chain, zhu2024large}. More recent efforts leverage multi-step reasoning, tool usage, and action execution \cite{yao2022react, hong2024datainterpreter, majumder2024discoverybench}, enabling LLMs to dynamically adjust to data characteristics and solve complex problems.

\subsection{Reproducibility in Data Science}



Prior work shows that reproducible data analysis requires not only access to code and data, but also clear documentation of the analyst’s reasoning and choices. \cite{goodman2016does, kale2019decision}. Unlike procedural tasks in software engineering or typical code generation settings \cite{hong2023metagpt, qian2023chatdev, islam2024mapcoder}, data analysis frequently involves multiple, \textit{equally justifiable} paths,
each potentially leading to \textit{different results} \cite{gelman2013garden, steegen2016increasing, simonsohn2020specification}. 
This analytical multiplicity, known as ``the garden of forking paths'', leads to substantial variation in outcomes even among experts analyzing the same data 
\cite{silberzahn2018many, bastiaansen2020time}. It is prevalent across domains 
\cite{botvinik2020variability, menkveld2024nonstandard}. 
and complicates efforts to assess an analysis’s validity from outputs alone. 

These prior findings highlight the importance of generating well-structured workflows that transparently communicate the analyst's plan, rationale, and execution steps, enabling independent verification and allowing reproducibility.

\subsection{Consistency of LLM Outputs}
Most existing work on reproducibility in LLMs centers on output consistency, i.e. whether an LLM produces the same output across repeated runs or prompt variations within similar conditions \cite{errica2024did, raj2025improving, ahn2025prompt}. For instance, \citet{wang2025assessing} systematically evaluates the consistency of LLM outputs across a range of financial tasks,
showing that the consistency rate varies across tasks.
Techniques such as self-consistency \citep{wang2022self, chen2023universal}, which aggregate final answers from multiple generations through majority voting, have been proposed to enhance answer accuracy.

However, these approaches focus on output consistency over repeated trials of one LLM rather than whether the underlying reasoning is explicit and complete enough for \textit{independent} reproduction -- a crucial aspect in data science, central to our work.



\section{Workflow Reproducibility} \label{sec:LLMRepr}


Reproducibility in data analysis depends on transparent workflows that clearly document key steps, such as data processing, modeling choices, and parameter settings, so others can reproduce results through independent implementation
\citep{davidson2008provenance, NAP25303}. We define {\it workflow reproducibility} as a property of the workflow reflecting how well it supports independent generation of code that faithfully implements the original analysis it describes. 
That is, it captures the alignment between what the workflow describes and what is actually implemented. 
 
To evaluate workflow reproducibility, we introduce an analyst-inspector framework for LLM-generated workflows: an LLM analyst produces a data science solution, and a separate LLM inspector attempts to independently reproduce the results using {\it only} the workflow and minimal context (e.g., dataset filepaths). 
Workflows that enable successful reproduction demonstrate strong fidelity between documentation and implementation.
This mirrors the reproducibility standard for publishing scientific research, where Methods sections must be sufficient for independent replication, and facilitates human-AI collaboration by shifting human oversight from labor-intensive manual code inspection to high-level reasoning about logic and assumptions made in the workflow (Figure \ref{fig:hci-pipeline}). 

\subsection{The Analyst-Inspector Framework}
\begin{figure}[t]
\centering
\includegraphics[width=\linewidth]{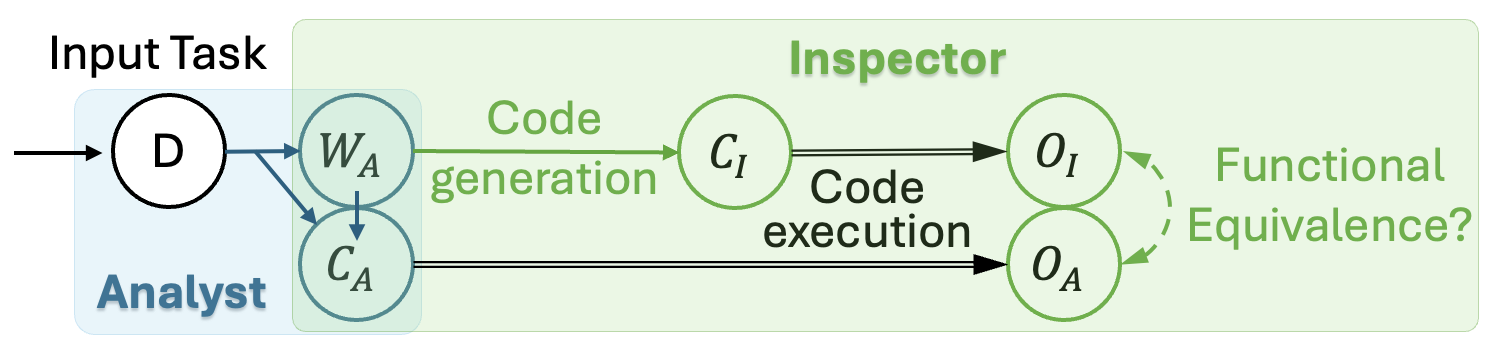}
  \caption{Analyst-Inspector framework for assessing LLM data analysis reproducibility. 
  }
\label{framework}
\end{figure}

Our formal reproducibility framework (Figure \ref{framework}) is defined as follows.
Let \( D \) represent a data science task, consisting of input data, contextual information, and a data science question (e.g. finding correlation). 
Let \( A \) be an AI analyst that generates analysis solutions according to probabilistic distribution \( f_A \) induced by the model's architecture and learned parameters, from which its statistical reasoning and code generation capabilities emerge.

When responding to task \( D \), analyst \( A \) produces a solution tuple $(W_A, C_A)$, where 

 $\bullet$ \( W_A \sim f_A(W \mid D)\) denotes the workflow, a structured summary encapsulating $A$'s reasoning and analysis plan for solving $D$. 
 
$\bullet$ \( C_A \sim f_A(C \mid D, W_A)\) denotes the corresponding code implementation following $W_A$.

To assess the reproducibility of $A$'s solution, we introduce an independent AI inspector \( I \) that evaluates whether the workflow \( W_A \) allows $I$ to reproduce the analysis conclusion by generating a new code implementation \( C_I \) from $W_A$, 
\vspace{-2mm}
\[
C_I \sim f_I(C \mid W_A),
\]  
where \( f_I \) is $I$'s probabilistic distribution for generating code given input text. We then define the criterion for a reproducible solution as follows.

\begin{definition*}
\label{def:reproducibility}

Given a task \( D \), workflow \( W_A \) is reproducible if and only if \( C_A \equiv C_I \), where \( \equiv \) holds if \( C_A \) and \( C_I \) produce the same result with respect to \( D \). 
If $W_A$ is reproducible, then its corresponding code implementation \( C_A \) is also reproducible making the analysis $(W_A, C_A)$ reproducible.
\end{definition*}

In our framework, \( C_A \equiv C_I \) denotes functional equivalence, meaning that the outcomes derived from their deterministic code execution, \( O_A=o(C_A) \) and \( O_I=o(C_I) \), are consistent, even if \( C_A \) and \( C_I \) differ textually. 

\subsection{Statistical Interpretations of Workflow Reproducibility}

Our reproducibility framework is grounded in the classical statistical principles of {\it sufficiency} and {\it completeness} \citep{casella2024statistical, lehmann1983theory}, which characterize how well a summary statistic (here, $W_A$) captures essential information for a given task. {\bf Sufficiency} ensures that the workflow contains all necessary information to reproduce the result; {\bf completeness} ensures that it excludes extraneous information that could distort replication.

Our definition of reproducibility (Section~\ref{def:reproducibility}) requires that the code distribution of an independent inspector given workflow matches that of the analyst given both data and workflow
\vspace{-2mm}
\[
f_A(C \mid D, W_A) = f_I(C \mid W_A).
\]
This condition enforces that \( W_A \) contains exactly the information necessary to generate functionally equivalent code -- no more, no less.

\textbf{Completeness} is reflected in the fact that irrelevant or ambiguous content in $W_A$ would cause the inspector’s code distribution to diverge from the analyst’s, leading to inconsistencies and violating the equality. A complete workflow avoids such derailment by eliminating non-essential detail.

{\bf Sufficiency} is expressed in two ways. First, when the inspector differs from the analyst (
 \(f_A \neq f_I\)), this condition ensures the workflow  generalizes across agents, guarding against idiosyncratic or agent-specific reasoning \citep{NAP25303}. Second, in the special case where the analyst and inspector are the same, the condition reduces to:
\vspace{-2mm}
\[
f_A(C \mid D, W_A) = f_A(C \mid W_A),
\]
which is equivalent to conditional independence \( C \perp D \mid W_A \). That is, the workflow sufficiently captures the relevant task context and renders the original data uninformative for code generation.

In this way, our formalization of reproducibility not only aligns with practical concerns around transparency and transferability, but also satisfies the foundational statistical properties of sufficiency and completeness.

\subsection{Reproducibility Enhancing Prompts} \label{sec:ReprEnhPrompt}
We propose two novel prompting strategies that explicitly aim to improve the reproducibility of LLM-generated data science workflows: 
\textbf{Reproducibility-of-Thought (RoT)} and \textbf{Reproducibility-Reflexion (RReflexion)}. \textbf{RoT} extends Chain-of-Thought (CoT) prompting \cite{wei2022chain} by adding an instruction that explicitly emphasizes both sufficiency and completeness:
``Make sure a person can replicate the action input by only looking at the workflow, and the action input reflects every step of the workflow.'' 
This encourages the LLM to generate workflows that are complete and confined by the implementation.
\textbf{RReflexion}, inspired by \citet{shinn2024reflexion}, 
introduces an iterative feedback mechanism: when a solution fails a reproducibility check by the inspector, the analyst is prompted to revise or regenerate the solution. This mirrors the human-in-the-loop process shown in Figure~\ref{fig:hci-pipeline} and simulates a cycle of feedback-driven refinement.


\section{Experiment Design} 
\subsection{Datasets} \label{sec:IntroData}
To evaluate the performance of LLMs in data analysis, we compiled a diverse set of 1,032 question-answer (QA) pairs (Table~\ref{tab:dataset}) sourced from three 
data science benchmarks: DiscoveryBench \citep{majumder2024discoverybench}, QRData \citep{liu2024llms}, and StatQA \citep{zhu2024large}. 
\textbf{1)} The DiscoveryBench benchmark comprises 264 tasks from real-world scientific problems and 903 synthetic tasks, covering 6 domains. We focused on the 239 real-world tasks in the `test' set.
\textbf{2)} The QRData benchmark tests advanced quantitative reasoning using 411 questions paired with data sheets from textbooks, online learning materials, and academic papers. For our evaluation, we removed 18 flawed multiple-choice questions as detailed in Appendix \ref{sec:Cat}.
\textbf{3)} The StatQA benchmark, designed for statistical analysis, contains 11,623 tasks spanning five categories: descriptive statistics (DS), correlation analysis (CA), contingency table tests (CTT), distribution compliance tests (DCT), and variance tests (VT). From its `mini-StatQA subset', consisting of 1,163 tasks, we randomly selected 80 tasks per category for balanced representation and to manage computational cost.

We used these benchmarks to assess LLMs' performance on various aspects of data analysis and statistical reasoning. Examples are provided in Appendix Table \ref{tab:CatExamples}. We classified all QA pairs into three categories: `Numerical', `Categorical', and `Textual' (Appendix \ref{sec:Cat}).

\subsection{LLM Models and Prompting Strategies} \label{sec:LLMAgents}
We evaluated five LLMs in combination with four prompting strategies. Specifically, LLMs included GPT-4o (GPT-4o-2024-11-20) \citep{openai2024gpt4}, Claude-3.5-sonnet (v2) \citep{TheC3}, o3-mini (o3-mini-2025-01-31) \citep{openai2025o3mini}, Llama-3.3-70B \citep{grattafiori2024llama3}, and DeepSeek-R1-70B \citep{guo2025deepseek}.

We evaluated four prompting strategies: two established baselines that do not explicitly enforce reproducibility, Chain-of-Thought (CoT) prompting \citep{wei2022chain} and ReAct \citep{yao2022react}, and our two reproducibility-enhancing methods, RoT and RReflexion (Section \ref{sec:ReprEnhPrompt}). CoT follows the approach introduced in \citet{wei2022chain}, using the line “let’s think step-by-step” to elicit reasoning. ReAct, a widely used agent-style prompting framework that interleaves reasoning with tool use \cite{yao2022react}, serves as our agent-based baseline for evaluating how well iterative decision-making supports reproducibility practices. 
As described in Section \ref{sec:ReprEnhPrompt}, RoT is CoT with an additional reproducibility-enhancing instruction, thus the comparison between RoT and CoT provides a direct assessment on the effect of reproducibility-enhancing instruction. RReflexion  extends CoT by incorporating inspector feedback from reproducibility check. In all the above prompts, we provide essential context, including detailed dataset descriptions and metadata, and any available relevant domain knowledge. All prompting strategies were equipped with the same code execution tool. Implementation constraints, including maximum number of LLM calls and code executions per sample for each prompting strategy are detailed in Table~\ref{tab:max_llm_code}.      

\subsection{Evaluation and Performance Analysis} \label{sec:EvalMetrics}
 
We assess each LLM-generated analysis using two binary metrics: accuracy and reproducibility. Accuracy measures whether the final answer matches the benchmark ground truth, e.g. `X is significantly larger than Y'. Reproducibility is assessed via our analyst-inspector framework (Section~\ref{sec:LLMRepr}). Solutions with inexecutable code are automatically marked as inaccurate and irreproducible. We use GPT-4o as the evaluator for both accuracy and reproducibility and validate its evaluations against those of a human expert on 350 samples, randomly drawn from each task category (Appendix \ref{sec:Alignment}). To assess framework robustness, we also test Claude-3.5-sonnet and o3-mini as the inspectors (Section \ref{Sec:Inspector}).

To analyze factors influencing performance, we fit a linear regression model: 
\vspace{-2mm}
\begin{equation*}
Y= \beta_0 + \beta_L LLM + \beta_p Prompt + \beta_d Task
\end{equation*}
where $Y$ is accuracy or reproducibility, and predictors include model, prompting strategy, and task type. Full specification and results are in Table \ref{tab:coef}. 

\subsection{Human Data Collection}
To contextualize the performance of LLMs, we establish a human analyst baseline using the DiscoveryBench dataset.  Two experienced postgraduate data analysts independently solved half of the tasks following the instructions in Appendix Table \ref{InstructionPrompt}. Each analyst spent approximately 120 hours developing comprehensive workflows and code solutions. We assessed these solutions using the methodology outlined in Section \ref{sec:EvalMetrics}.

\begin{figure}[htbp]
  \includegraphics[width=\linewidth]{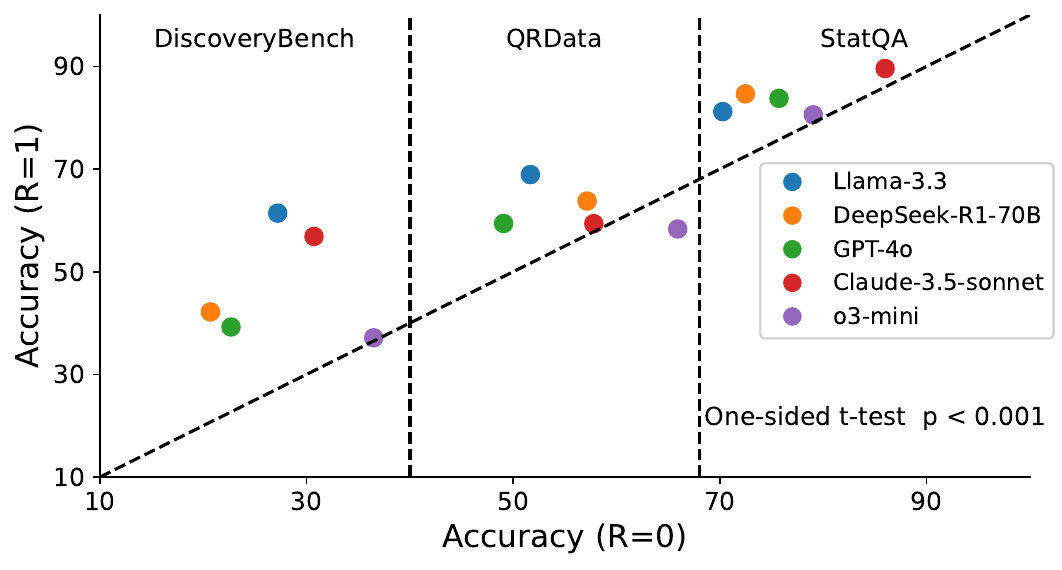} 
  \caption {Accuracy comparison between reproducible ($R=1$) and irreproducible ($R=0$) solutions across LLMs and datasets using CoT prompting. Non-executables are excluded from the $R=0$ group. The x- and y-axes show the proportion of accurate solutions in each group. The diagonal line indicates equal accuracy between the two groups. A one-sided t-test evaluates whether reproducible solutions are significantly more accurate.}
  \label{AccAtRepr}
\end{figure}

\section{Results}

\begin{figure*}[htbp]
  \includegraphics[width=\linewidth]{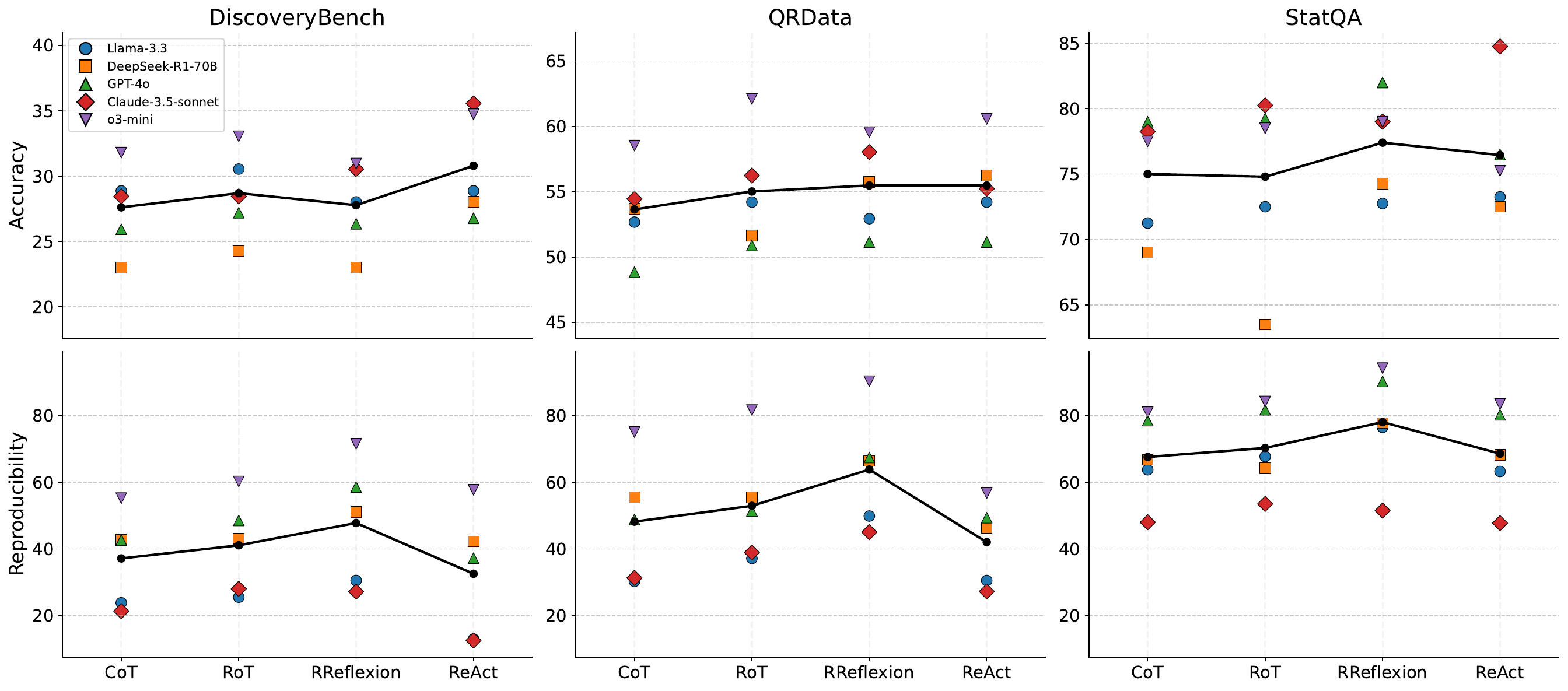} 
  \caption {Accuracy and reproducibility of LLMs across datasets for different prompting strategies.  
  Solid line: the average metric scores calculated from different LLMs for each prompting strategy.
}
\label{AccRepr}
\end{figure*}

\subsection{Reproducible Analyses are more Accurate}
To evaluate the relationship between reproducibility and correctness of the LLM-generated analyses, we compared accuracy between reproducible and irreproducible solutions for each LLM-dataset combination. As shown in Figure~\ref{AccAtRepr}, reproducible solutions were significantly more accurate than irreproducible ones (mean 64.4\% vs. 53.5\%; one-sided paired t-test, p < 0.001). 
This pattern held consistently across prompting strategies (Figure \ref{AccAtReprAgents}), suggesting that workflows detailed enough to support independent replication are more likely to reflect sound analytical reasoning. 

\subsection{Impact of Prompting Strategies}

Prompting strategies that explicitly enforce reproducibility—RoT and RReflexion—substantially improved both reproducibility and accuracy across models and datasets. Compared to the baseline CoT strategy, RoT increased reproducibility rates most of the time. For instance, on DiscoveryBench with GPT‑4o, reproducibility rose from 42.68\% (CoT) to 48.54\% (RoT), and RReflexion further increased it to 58.58\% (Table \ref{tab:discoverybench}). Similarly, o3‑mini improved from 55.23\% (CoT) to 60.25\% (RoT), peaking at 71.55\% with RReflexion. Regression analysis confirmed that RReflexion had a statistically significant positive effect on reproducibility across all datasets (Table~\ref{tab:coef}).

These improvements in reproducibility were accompanied by gains in analytical correctness. RoT outperformed its baseline CoT counterpart across nearly all settings (Tables~\ref{tab:discoverybench} and~\ref{tab:qrdata-statqa}). RReflexion yielded a significant (p = 0.007, paired t-test) 4.15\% average accuracy gain on reanalyzed samples over CoT. These findings further underscore reproducibility as a practical and reliable proxy for analytical quality.

Notably, the accuracy improvements were more pronounced on benchmarks with higher accuracy, such as QRData and StatQA. For example, on QRData, GPT‑4o's accuracy increased from 48.85\% (CoT) to 50.89\% (RoT) and 51.15\% (RReflexion), while o3‑mini improved from 58.52\% (CoT) to 62.09\% under RoT. In contrast, gains were more modest on DiscoveryBench. These results imply that, although reproducibility-focused prompts enhance alignment between reasoning and execution, they alone aren’t enough to address the harder statistical challenges posed by more complex tasks.

We also assessed ReAct, an agent-style prompting framework that interleaves workflow reasoning and code execution. Consistent with prior work \citep{yao2022react}, ReAct often improved accuracy over CoT, particularly for Claude-3.5-sonnet and DeepSeek-R1-70B (Figure \ref{AccRepr}). However, ReAct sometimes reduced reproducibility. For instance, Llama‑3.3’s reproducibility on DiscoveryBench dropped from 23.85\% (CoT) to 12.97\% (ReAct). This decline appears to result from incomplete workflows, as models occasionally referenced prior steps without rearticulating them, despite explicit instructions to produce complete workflows (Appendix \ref{sec:PromptTemp}). Nonetheless, ReAct, by leveraging iterative tool feedback, markedly reduced execution errors and delivered the highest proportion of executable code (Figure~\ref{CodeNotRun}).

\subsection{Comparison Across LLMs}
Comparing performance across the three benchmark datasets (Appendix \ref{sec:AppendixA}), all LLMs performed best on StatQA (mean accuracy 75.91\%), followed by QRData (54.9\%), with the lowest performance on DiscoveryBench (28.72\%). Variations in reproducibility (Figure \ref{AccRepr}) were less striking, but followed the same trend. This pattern reflected the varying level of task complexity across the datasets (easiest to hardest: StatQA, QRData, DiscoveryBench).

As analysts, different LLMs demonstrate substantial variation in analytical correctness, workflow reproducibility, and code reliability. Among them, o3-mini achieved the strongest overall performance: it topped accuracy on DiscoveryBench (32.64\%) and QRData (60.18\%), achieved the highest reproducibility across all datasets (up to 94.25\%), and generated the most executable code (Tables \ref{tab:discoverybench} and \ref{tab:qrdata-statqa}). Claude-3.5-sonnet often achieved high accuracy, including the best score on StatQA (80.56\%), but consistently showed the lowest reproducibility, indicating that its workflows often do not support reliable reproduction. 
GPT-4o exhibited inconsistent performance across benchmarks, ranking among the best on StatQA, lowest on QRData, and mid-level on DiscoveryBench. Llama-3.3 showed steady mid-tier performance across all metrics. DeepSeek-R1-70B performed the worst overall, exhibiting low accuracy, and the lowest rate of executable code.


\subsection{Performance Across Task Types}
\begin{figure}[htbp]
  \includegraphics[width=\linewidth]{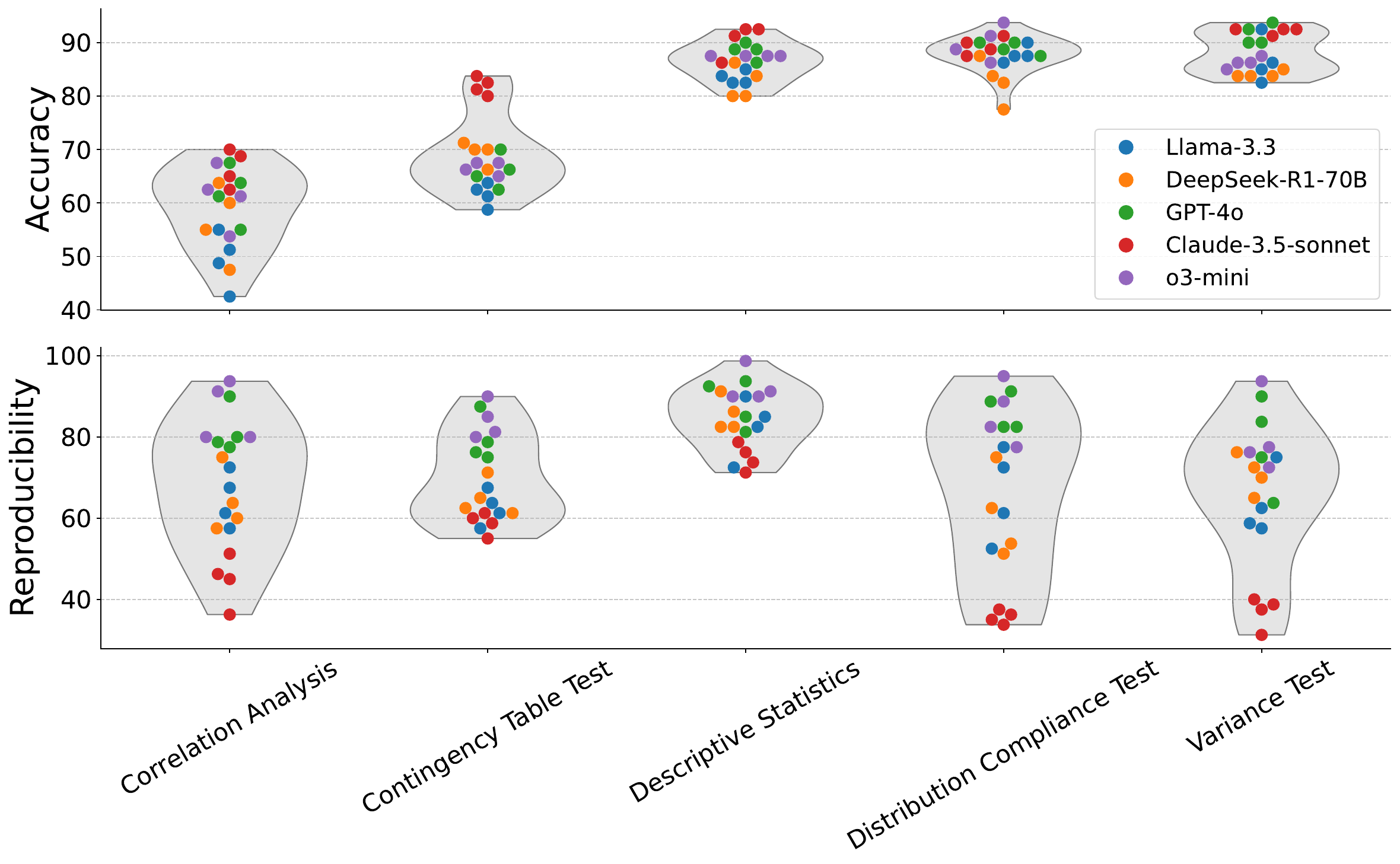} 
  \caption {Accuracy and reproducibility of LLMs across different statistical question categories in StatQA.
 }
 \label{StatQACtgr}
\end{figure}

To evaluate how LLMs perform across different types of analysis tasks, we examined results by statistical question category within the StatQA dataset (Figure \ref{StatQACtgr}, category details in Section \ref{sec:IntroData}). Descriptive Statistics consistently yielded the highest accuracy (80.00–92.50\%) and reproducibility (71.25–98.75\%), with o3-mini performing particularly well (90.00–98.75\%). DCT and VT tasks also had high accuracy (77.50–93.75\%), though reproducibility differed widely across models, ranging from 72.50–95.00\% for o3-mini to 31.25–40.00\% for Claude-3.5-sonnet. In contrast, tasks involving CTT and CA were more challenging, with accuracy ranged from 42.50–83.75\%. Reproducibility also varied widely: Claude-3.5-sonnet ranged between 36.25–51.25\%, while o3-mini maintained consistently high reproducibility (80.00–93.75\%).

We also grouped the 1,032 QA tasks into three data types—numerical, categorical, and textual—to assess how data type affects model performance (Figure \ref{Ctgr}). Regression analysis (Table~\ref{tab:coef}) revealed that while categorical and textual tasks had higher accuracy than numerical ones, they were significantly less reproducible. 

\subsection{Error Analysis for Irreproducibility}
\begin{figure}[htbp]
  \includegraphics[width=0.95\linewidth]{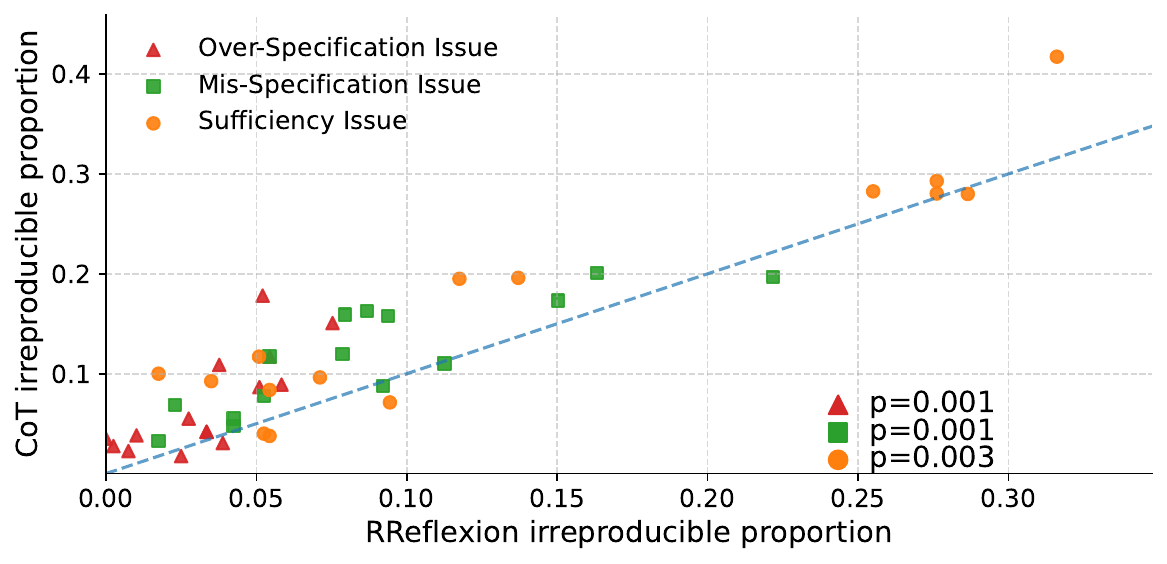} 
  \caption {Comparison between CoT and RReflexion on different types of irreproducibility. P values are from two-sample one-sided t-tests for RReflexion $<$ CoT.
  }
  \label{error_analysis_comparison}
\end{figure}

To better understand the causes of irreproducibility in LLM-generated workflows, we categorized irreproducible workflows into three distinct types according to their misalignment with the Analyst's code implementation (Appendix~\ref{Sec:AppErrorAnalysis}):  {\bf Sufficiency Issue}: Missing essential steps or details present in the code. {\bf Mis-Specification Issue}: All steps are included but some diverge from the code's logic — e.g., wrong statistical methods or misdefined variables. {\bf Over-Specification Issue}: Includes unnecessary steps, constraints or assumptions not in the code, introducing ambiguity. Demonstration examples are provided in Figure ~\ref{error_example}.

Among these categories, sufficiency issues (49.94\%) were most common (Figure \ref{Error_Analysis}), especially in outputs from Claude-3.5-sonnet and Llama-3.3. Although less frequent, mis-specification (33.02\%) and over-specification (17.04\%) are more damaging, introducing incorrect  procedures or a false sense of rigor through non-implemented steps.
Notably, reproducibility-enhancing prompts, especially RReflexion, significantly reduced all error types (Figure \ref{error_analysis_comparison}), highlighting 
 the value of inspector feedback in improving workflow clarity and fidelity. RoT also yielded significant gains (Figure \ref{error_analysis_CoT_vs_RoT}).

\subsection{Robustness to Inspector Choice} \label{Sec:Inspector}
\begin{figure}[t]
\centering
\includegraphics[width=1\linewidth]{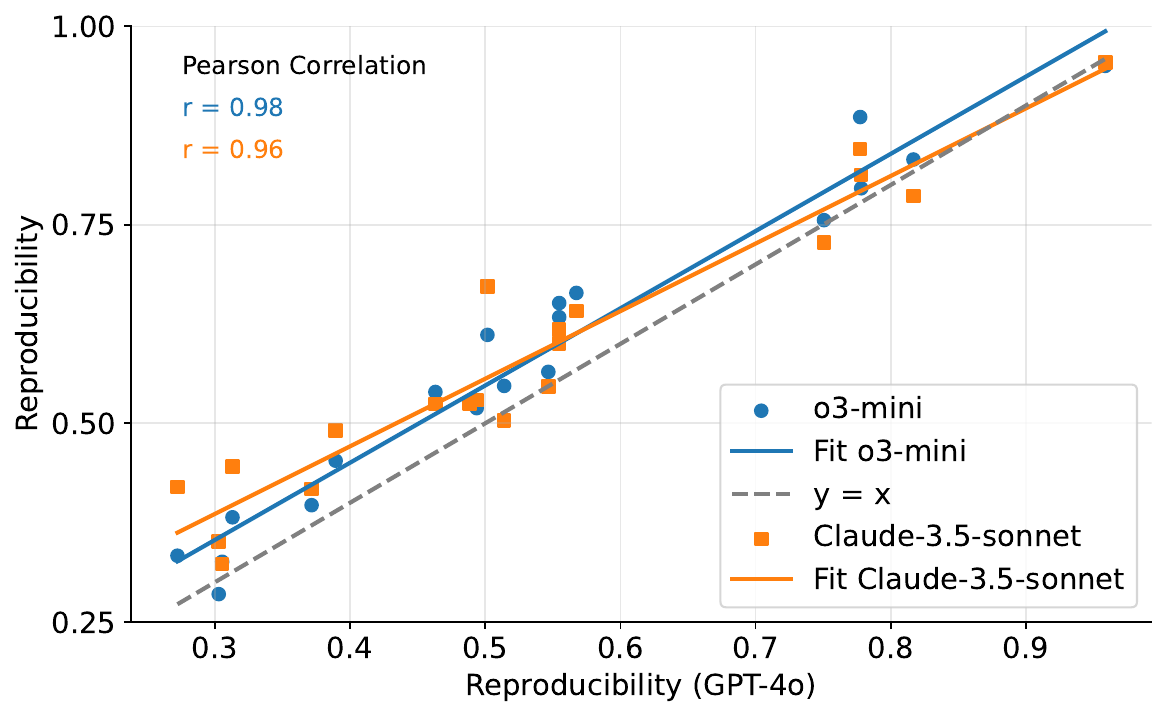}
  \caption{Agreement in reproducibility assessments among GPT-4o, o3-mini, and Claude-3.5-sonnet inspector LLMs on the QRData dataset.}
\label{inspector_together}
\end{figure}


To assess the sensitivity of our framework to inspector choice, we re-evaluated reproducibility using two additional LLMs, Claude-3.5-sonnet and o3-mini, on the QRData benchmark, which features moderate analytical complexity. Their reproducibility scores were strongly correlated with those from GPT-4o across all analyst models and prompting strategies (Figure \ref{inspector_together}), with only minor score increases. We also assessed self-evaluation bias. Claude-3.5-sonnet rated its own outputs consistently higher than other inspectors did (Figure \ref{inspector}), suggesting potential bias. GPT-4o and o3-mini showed no such tendency.





These findings support the robustness of our framework across inspector choices and identify GPT-4o and o3-mini as more consistent, unbiased inspectors.

\subsection{Human vs LLM Comparison}
Comparing human expert solutions with LLM solutions on DiscoveryBench, we observed a notable performance gap in overall accuracy (Table \ref{tab:discoverybench} and Appendix Table \ref{LLMvsHuman}). Human analysts achieved an overall accuracy of 66.53\% and a reproducibility rate of 66.53\%, albeit at the cost of approximately 240 hours of manual effort. LLMs, while less accurate overall (23.01–35.56\%), demonstrated competitive reproducibility under certain configurations. Notably, o3-mini with RReflexion (71.6\%) surpassed human reproducibility. These results suggest that future enhancements in prompt design and reproducibility practices could help narrow the gap between automated and manual approaches.

\section{Conclusion}
In this work, we introduce AIRepr, an analyst-inspector framework designed to rigorously evaluate and enhance the reproducibility of LLM-generated data analysis workflows. Our systematic evaluation demonstrates that enforcing reproducibility criteria on workflows can improve the accuracy of analyses, underscoring the value of explicit and transparent reasoning in AI-driven data science. 


Embedding reproducibility principles through our RoT and RReflexion strategies substantially boosts both reproducibility and accuracy across five LLMs, outperforming standard CoT prompting, and demonstrating that reproducibility is not only a desirable property but a practical means of improving performance.
Error analysis further revealed that inspector feedback in our RoT and RReflexion strategies effectively reduces common causes of irreproducibility, such as omissions, mis-specifications, and unnecessary steps that introduce ambiguity.

We also find that AIRepr is robust to the choice of Inspector model. While some models, like Claude-3.5-sonnet, may overrate their own outputs, others such as GPT-4o and o3-mini provide more consistent and unbiased evaluations, making them more suitable for inspection.

The AIRepr framework is broadly applicable to LLM-based systems that generate both workflows and executable code. By introducing an Inspector to independently verify Analyst outputs, users can implement a lightweight reproducibility check before relying on results. This approach may extend to real-world tools with similar structures, such as Microsoft Copilot Analyst or Perplexity Lab, as a modular addition for improving trust and reliability in automated data analysis.

Overall, AIRepr promotes more transparent, reliable, and efficient human-AI collaboration, enhancing the integrity and usability of automated data analysis.


\section{Acknowledgements}
This work is partially supported by NIH R01 GM109453 to QL. 

\section{Limitations}
Our study has several limitations that suggest directions for future work.

First, while we explored a wide range of prompting strategies to improve reproducibility, we did not examine the potential of fine-tuning LLMs on task-specific data or reproducibility-oriented objectives. Fine-tuning could further enhance the quality and consistency of generated workflows.


Second, although we use an independent Inspector model to assess reproducibility and incorporate human verification, shared training data and alignment objectives across LLMs may still introduce agreement artifacts. This could lead to inflated reproducibility scores even when the outputs are not independently correct. We acknowledge this risk and note that future extensions to our framework could include using a diverse ensemble of Inspector models or more stringent validation strategies to reduce such false positives.

Third, the benchmark datasets used in our evaluation span diverse domains and task types, but they may still fall short of representing the full complexity, ambiguity, and domain-specific nuance found in real-world data analysis workflows.

Finally, we acknowledge that many data science tasks are inherently open-ended and may admit multiple valid solutions. Our evaluation of accuracy relies on comparison to predefined benchmark answers, which may not capture the full range of defensible outcomes. 

\section{Ethics Statement}

\paragraph{Existing Benchmark Licenses.} 
Our work builds upon publicly available datasets and models, ensuring compliance with their respective licenses. Below, we detail the licensing terms for the benchmarks we used and our own contributions.
\begin{itemize}
    \item \textbf{DiscoveryBench} \cite{majumder2024discoverybench}: Licensed under ODC-BY, permitting redistribution and modification with proper attribution.
    \item \textbf{QRData} \cite{liu2024llms}: Licensed under CC BY-NC 4.0, allowing non-commercial use with attribution.
    \item \textbf{StatQA} \citep{zhu2024large}: Licensed under GPL-3.0, requiring derivative works to adopt the same license.
\end{itemize}
Our use of these datasets adheres to the terms specified by their respective licenses, and we use them strictly for research purposes.

\paragraph{Codebase License.} We release our code under the MIT License, granting users permission to use, modify, and distribute the code with proper attribution. This choice aligns with our goal of advancing reliability of LLM-generated data analysis and reproducibility of scientific research.

\paragraph{Potential Risks.}
Our evaluation indicates that current LLMs are not yet fully reliable in generating reproducible and accurate data analyses. We suggest that users carefully review any LLM-generated solutions before using them to automate data analysis tasks. Additionally, while reproducible analyses are more likely to be correct, it is possible that both the analyst and inspector agents are incorrect in consistent ways
due to shared model biases or alignment artifacts. To mitigate such false negatives, our framework supports modular extensions such as incorporating multiple inspectors to perform re-audits, flag disagreements, or apply stricter consensus mechanisms (e.g., majority voting). Thus, the AIRepr framework is designed as a human-in-the-loop system, as detailed in our motivation and Figure \ref{fig:hci-pipeline}, with the goal of reducing human workload while increasing trust in LLM-assisted data analysis workflows.

\bibliography{main}

\appendix

\section{Data Cleaning and QA Types} \label{sec:Cat}
The questions removed from the QRData benchmark were 18 causal relationship questions with one of the following two structures:

``Question: Which cause-and-effect relationship is more likely?

Answer options:

A: [item] causes [item] 

B: [item] causes [item] 

C:  The causal relation is double sided between [item] and [item]

D: No causal relationship exists''

``
Question: Which cause-and-effect relationship is more likely?

Answer options:

A: [item] causes [item] 

B: [item] causes [item] 

C: No causal relationship exists''

where all [item]'s are the same, rendering the answer options meaningless and duplicated.

`Numerical' responses provide quantitative values; `Categorical' responses consist of discrete labels or classifications; `Textual' responses involve free-form language that delivers narrative descriptions, interpretations, or explanations. Appendix Table \ref{tab:CatExamples} provides examples for each QA type. For StatQA, if the hypothesis test responses contradict each other, the QA type is classified as `Textual'.

\section{Description of Results Tables and Figures} \label{sec:AppendixA}
\begin{figure*}[htbp]
  \includegraphics[width=\linewidth]{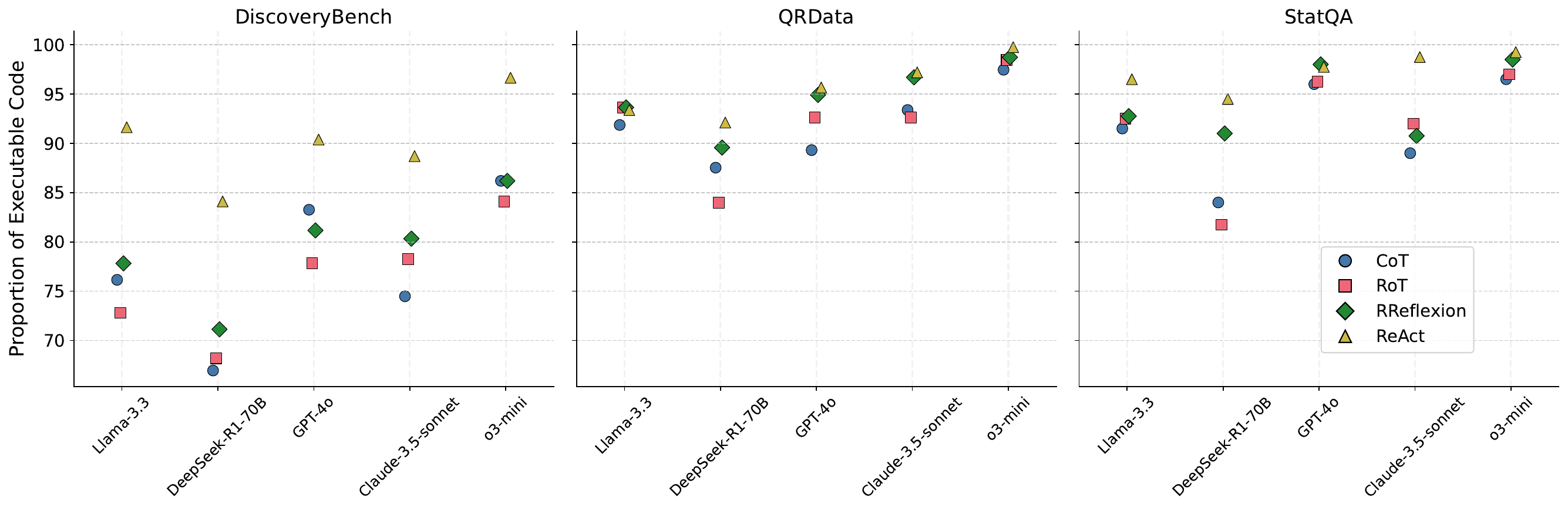} 
  \caption {Proportion of executable code across datasets for various LLMs and agents. 
  }
  \label{CodeNotRun}
\end{figure*}

\begin{figure*}[htbp]
  \includegraphics[width=\linewidth]{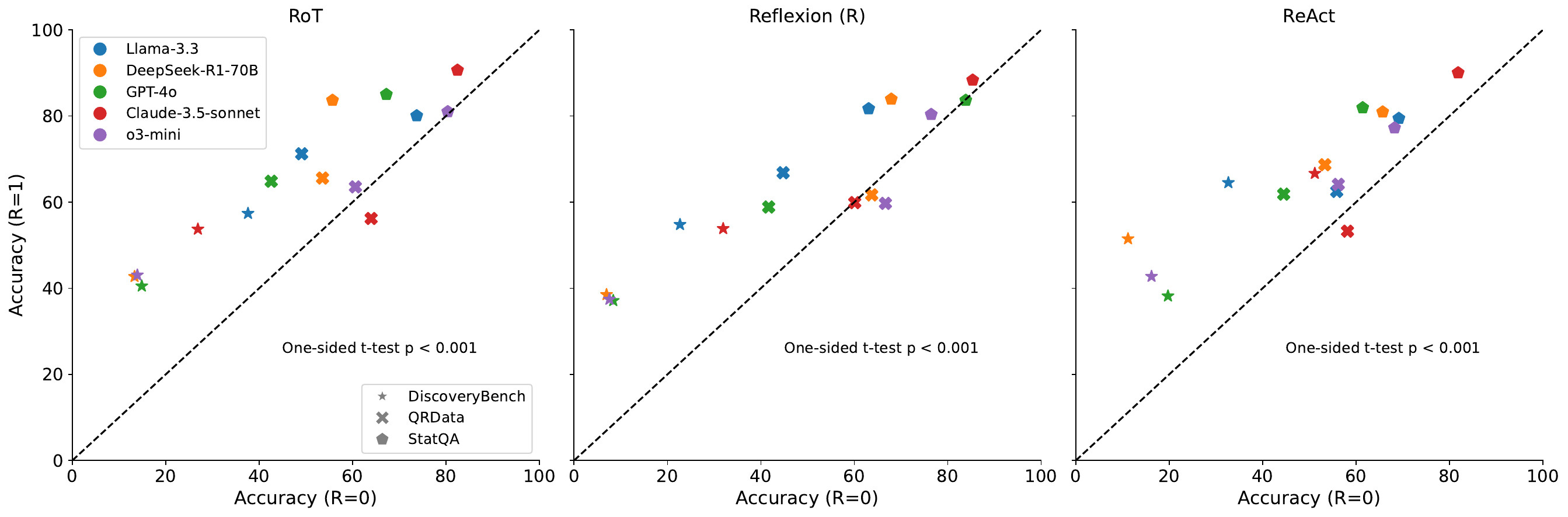} 
  \caption {Comparison of accuracy across datasets for RoT, RReflexion, and ReAct prompting strategies using LLM models at varying reproducibility levels. For accuracy calculations at R=0, samples containing inexecutable code are excluded. The p values of paired one-sided t-tests are all less than 0.001.}
  \label{AccAtReprAgents}
\end{figure*}

Appendix Tables \ref{tab:discoverybench} and \ref{tab:qrdata-statqa} were used to generate Figures \ref{AccRepr} and \ref{AccAtRepr}, along with Appendix Figures \ref{CodeNotRun} and \ref{AccAtReprAgents}. Appendix Tables \ref{CACTT}, \ref{DSDCT}, and \ref{VT} were used to generate Figure \ref{StatQACtgr}. Appendix Tables \ref{CACTT}, \ref{DSDCT}, and \ref{VT} showed the results of LLMs on different categories of StatQA tasks. Appendix Tables \ref{CatNum} and \ref{TXT} showed the results of LLMs on Numerical, Categorical and Textural QA types.

\section{Prompt Templates} \label{sec:PromptTemp}
Templates for different prompting strategies are provided in Appendix Tables \ref{CoTPrompt}, \ref{ReActPrompt}, \ref{ContinueReActPrompt}, and \ref{Continue2ReActPrompt}:
\begin{itemize}
\setlength{\itemindent}{0em}
    \item Appendix Table \ref{CoTPrompt}: CoT, RoT, and RReflexion templates
    \item Appendix Tables \ref{ReActPrompt}, \ref{ContinueReActPrompt}, \ref{Continue2ReActPrompt}: ReAct template
\end{itemize}

\section{Evaluation} \label{sec:EvalPrompt}
To compute accuracy, analysis correctness was determined by comparing the LLM analysis conclusions to the benchmark ground truth. For numerical answers, the LLM conclusion agrees with the ground truth if the deviation between LLM conclusion and the ground truth is within a predefined error threshold (Table \ref{AccuracyPrompt}). As the majority of LLM conclusions and ground truth answers are presented in natural language, we employ an evaluator LLM (GPT-4o-2024-11-20) to assess agreement with the ground truth. Reproducibility is evaluated via the procedure detailed in Section \ref{sec:LLMRepr}. Prompts for the answer agreement evaluation and reproducibility inspection are detailed in the below prompts. 

For Llama-3.3, GPT-4o, Claude-3.5-sonnet, and o3-mini, we used a regex parser to extract the workflow, code, and conclusion from the LLM output. For DeepSeek-R1-70B, the parser extracted the code and conclusion, and all reasoning text preceding the code was treated as the workflow. For CoT and RoT, which generate a single iteration of workflow and code, reproducibility was assessed on that output. For RReflexion and ReAct, which are iterative, only the final workflow and code were evaluated. 

Evaluation prompt templates are provided in Appendix Tables \ref{AccuracyPrompt}, \ref{WorkflowCodePrompt}, \ref{ConclusionPrompt}, and \ref{ReprPrompt}:
\begin{itemize}
\setlength{\itemindent}{0em}
    \item Appendix Table \ref{AccuracyPrompt}: Accuracy assessment 
    \item Appendix Table \ref{WorkflowCodePrompt}: Workflow-to-code conversion
    \item Appendix Table \ref{ConclusionPrompt}: Conclusion extraction from code execution output
    \item Appendix Table \ref{ReprPrompt}: Reproducibility assessment
\end{itemize}

\section{Assessing Alignment of LLM Evaluator and Human Judgments} \label{sec:Alignment}
To validate the reliability of our automated evaluation framework, we conducted a systematic human validation study comparing manual scoring against LLM-generated scores.

For accuracy assessment, we randomly selected 25 examples from each QA type (numerical, categorical, and textual) across all three datasets (DiscoveryBench, QRData, and StatQA) from the GPT-4o CoT results. A human expert then independently assessed the accuracy of the LLM's predicted answers. Overall, among 200 examples, we found a 98.5\% agreement between the human evaluations and the LLM scoring, underscoring the high reliability of our automated accuracy assessment.

For reproducibility assessment, we randomly selected 50 examples from each dataset in the GPT-4o CoT results, resulting in a total of 150 samples. A human expert manually evaluated these examples using our established criteria in Section \ref{sec:LLMRepr}. The evaluation showed a 98.7\% agreement between the human assessments and the LLM scoring, confirming that our automated reproducibility assessment closely aligned with human judgment.

These high consistency rates (98.5\% for accuracy and 98.7\% for reproducibility) confirmed the reliability of our automated evaluation and demonstrated its suitability for large-scale assessments of LLM performance in data science tasks.

\section{Irreproducibility Error analysis} \label{Sec:AppErrorAnalysis}
To diagnose why certain AI-generated workflows fail our reproducibility check, we classify each irreproducible sample into one of three mutually exclusive error categories: Workflow Sufficiency Issues, Workflow Mis-Specification Issues, and Workflow Over-Specification Issues.

We automate the category assignment using o3-mini (high) with a dedicated prompt (Appendix Table~\ref{WorkflowReproPrompt}). For each irreproducible sample, we provide: 1) the analyst's generated workflow, 2) the analyst's code implementation, 3) the workflow-converted code by inspector, and 4) the reason why the inspector thinks the sample is irreproducible. Then we provide the definitions of the three irreproducibility categories with examples to clarify the failure reason and guide the categories selection.  

\section{Human Experts}

The two human expert data analysts who provided workflow and code solutions in this study were postgraduates (PhD, MS) with extensive experience in data analysis and fluency in English. They were recruited through the authors' personal connections. They were not paid, but instead credited as co-authors of this work. They were provided the instructions in Appendix Table \ref{InstructionPrompt} and were informed how their analysis solutions would be used.

\section{Computational Resources and Experimental Setup}

We conducted our experiments on a mix of locally hosted (open-source) and API-accessed (closed-source) LLMs. The locally hosted models included Llama-3.3-70B and DeepSeek-R1-70B, both of which were run on a cluster of four NVIDIA A6000 GPUs. The total computational budget was approximately 200 GPU hours for Llama-3.3-70B and 300 GPU hours for DeepSeek-R1-70B.  

For consistency, we set temperature to 0 for all models except o3-mini, which required a temperature of 1 to ensure compatibility with the LangChain AzureChatOpenAI package. The hyperparameters for each analyst model were as follows:

\begin{itemize}
    \item \textbf{Llama-3.3-70B}: \texttt{repetition\_penalty = 1.18}, \texttt{num\_ctx = 8192}, \texttt{num\_predict = 2048}
    \item \textbf{DeepSeek-R1-70B}: \texttt{num\_ctx = 8196 \(\times\) 2}, \texttt{num\_predict = 6000}
    \item \textbf{GPT-4o}:  \texttt{max\_tokens = 2048}
    \item \textbf{Claude 3.5 Sonnet}: \texttt{max\_tokens = 2048}
    \item \textbf{o3-mini}:  \texttt{reasoning\_effort = low}, \texttt{max\_completion\_tokens = 4000}
\end{itemize}

We did not perform extensive hyperparameter tuning beyond selecting reasonable values based on prior literature and preliminary trials. Instead, we focused on evaluating model performance on structured data analysis tasks under fixed conditions to ensure a fair comparison across LLMs.

The LLM snapshots we used for the OpenAI models were 2025-01-31 for o3-mini and 2024-11-20 for GPT-4o.
The Azure API versions used were 2024-12-01-preview for o3-mini and 2024-10-01-preview for GPT-4o.

\section{Use of AI Assistants}

We used GitHub Copilot for code completion and Anthropic's Claude web interface to assist with debugging. Additionally, we used OpenAI's ChatGPT web interface to refine the clarity and smoothness of our writing. 

All AI-generated content, whether in code or text, was carefully reviewed, edited, and validated by the authors to ensure accuracy and alignment with our research objectives. No AI-generated content was used without human verification, and all final decisions regarding implementation and manuscript writing remained with the authors.

\begin{table*}
  \centering
  \begin{tabular}{p{0.95\textwidth}}
    \hline
Workflow: \\
1. Load the CSV dataset 'Fruits\_and\_Vegetables\_Prices\_Dataset.csv' into a pandas DataFrame.\\
2. Use a linear regression model (using statsmodels) where \texttt{newyorkretail} is the dependent variable and both \texttt{farmprice} and \texttt{averagespread} are independent variables.\\
3. Print the summary of the fitted regression model.\\[1ex]

Code 1: \\
\begin{verbatim}
import pandas as pd
import statsmodels.api as sm

# Load dataset
df = pd.read_csv('Fruits_and_Vegetables_Prices_Dataset.csv')

# Define independent variables and dependent variable
X = df[['farmprice', 'averagespread']]
y = df['newyorkretail']

# Fit the Ordinary Least Squares (OLS) regression model
X = sm.add_constant(X)
model = sm.OLS(y, X).fit()

# Display the summary of the regression model
print(model.summary())
\end{verbatim}\\[1ex]

Code 2: \\
\begin{verbatim}
import pandas as pd
import statsmodels.api as sm

# Step 1: Load the CSV dataset into a pandas DataFrame
dataset_path = 'Fruits_and_Vegetables_Prices_Dataset.csv'
data = pd.read_csv(dataset_path)

# Step 2: Define the dependent and independent variables
X = data[['farmprice', 'averagespread']]  # Independent variables
y = data['newyorkretail']  # Dependent variable

# Add a constant to the independent variables for the intercept
X = sm.add_constant(X)

# Fit the linear regression model using statsmodels
model = sm.OLS(y, X).fit()

# Step 3: Print the summary of the regression model to view the coefficients
print(model.summary())
\end{verbatim}\\
    \hline
  \end{tabular}
  \caption{\label{WorkflowCodeExample}
    An example where a reproducible workflow enables the functional equivalent reproduction of Code 1 via Code 2.
  }
\end{table*}

\begin{table*}
  \centering
  \begin{tabular}{lccc|c}
    \hline
    \textbf{Dataset} & \textbf{Num} & \textbf{Cat} & \textbf{Txt} & \textbf{Total} \\
    \hline
    DiscoveryBench   & 90                 & 86                   & 63               & 239           \\
    QRData           & 163                & 230                  & 0                & 393           \\
    StatQA           & 75                 & 249                  & 76               & 400           \\
    \hline
    \textbf{Total}   & 328                & 565                  & 139              & 1032          \\
    \hline
  \end{tabular}
  \caption{
  Distributions of QA pairs by dataset. 
  }
  \label{tab:dataset}
\end{table*}

\begin{table*}[htbp]
    \centering
    \begin{tabular}{@{}p{0.17\linewidth} p{0.35\linewidth} p{0.25\linewidth} p{0.11\linewidth}@{}}
        \toprule
        \textbf{Dataset} & \textbf{Question} & \textbf{Conclusion} & \textbf{Type} \\ \midrule
        DiscoveryBench & What is the relationship of amber finds and number of monuments between 3400-3000 BCE? & Between 3400-3000 BCE, there is a high number of amber finds and a large number of monuments. & Textual \\[1ex]
        \midrule
        QRData & Which cause-and-effect relationship is more likely? A. Lumbago causes R S1 radiculopathy B. R S1 radiculopathy causes Lumbago & B & Categorical \\[1ex]
        \midrule
        StatQA & What is the kurtosis of the distribution of the variable representing Base Special Defense? & 2.39175 & Numerical \\ \bottomrule
    \end{tabular}
    \caption{Examples of different QA types.}
    \label{tab:CatExamples}
\end{table*}

\begin{table*}[htbp]
\centering
\begin{tabular}{lcc}
\toprule
Prompt           & \begin{tabular}[c]{@{}c@{}}Max LLM\\ Call\end{tabular} & \begin{tabular}[c]{@{}r@{}}Max Code\\ Execution\end{tabular}  \\
\midrule
CoT            & 2            & 1                  \\
RoT            & 2            & 1                  \\
RReflexion      & 3            & 1                  \\
ReAct          & 4            & 3                  \\
\bottomrule
\end{tabular}
\caption{In data analysis, following code execution, the same LLM was reused solely to generate an answer to the analysis question. Maximum number of LLM calls and code executions across different prompting strategies.}
\label{tab:max_llm_code}
\end{table*}

\begin{table*}
  \centering
  \begin{tabular}{p{0.95\textwidth}}
    \hline
Could you help work out the workflow (analysis steps) and code solutions on DiscoveryBench questions?\\

Below are the instructions:\\
 
To prepare solutions for given questions in a standardized format. Each solution should include:\\
1. A workflow explaining the steps and logic of the code.\\
2. Python code that computes the answer.\\[1ex]
 
Additional Requirements:\\
1.  The code should be written in a way that, when provided to another person, they can write a workflow with the same level of detail and functionality as your workflow.\\
2. The workflow should be detailed and clear enough that, when provided to another person, they can write functionally identical code solely based on the workflow.\\
    \hline
  \end{tabular}
  \caption{Instructions given to human experts for solving data analysis problems in DiscoveryBench dataset.}
  \label{InstructionPrompt}
\end{table*}

\begin{table*}[htbp]
\centering
\begin{tabular}{ll|rrr|rr}
\toprule
Prompt           & Model                & \begin{tabular}[c]{@{}r@{}}Overall\\ Accuracy\end{tabular} & \begin{tabular}[c]{@{}r@{}}Overall\\ Reproducibility\end{tabular} & \begin{tabular}[c]{@{}r@{}}Inexecutable\\ Code\end{tabular} & \begin{tabular}[c]{@{}r@{}}Accuracy\\ (R=1)\end{tabular} & \begin{tabular}[c]{@{}r@{}}Accuracy\\ (R=0)\end{tabular} \\
\midrule
\multirow{5}{*}{CoT} 
    & Llama-3.3         & 28.87  & 23.85  & 23.85 & 61.4  & 27.2 \\
    & DeepSeek-R1-70B   & 23.01  & 42.68  & 33.05 & 42.16 & 20.69 \\
    & GPT-4o            & 25.94  & 42.68  & 16.74 & 39.22 & 22.68 \\
    & Claude-3.5-sonnet & 28.45  & 21.34  & 25.52 & 56.86 & 30.71 \\
    & o3-mini           & 31.80  & 55.23  & 13.81 & 37.12 & 36.49 \\
\cmidrule{1-7}
\multirow{5}{*}{RoT} 
    & Llama-3.3         & 30.54  & 25.52  & 27.2  & 57.38 & 33.63 \\
    & DeepSeek-R1-70B   & 24.27  & 43.10  & 31.80 & 42.72 & 23.33 \\
    & GPT-4o            & 27.2   & 48.54  & 22.18 & 40.52 & 25.71 \\
    & Claude-3.5-sonnet & 28.45  & 28.03  & 21.76 & 53.73 & 26.67 \\
    & o3-mini           & 33.05  & 60.25  & 15.90 & 43.06 & 29.82 \\
\cmidrule{1-7}
\multirow{5}{*}{RRefl.} 
    & Llama-3.3         & 28.03  & 30.54  & 22.18 & 54.79 & 23.89 \\
    & DeepSeek-R1-70B   & 23.01  & 51.05  & 28.87 & 38.52 & 16.67 \\
    & GPT-4o            & 26.36  & 58.58  & 18.83 & 37.14 & 20.37 \\
    & Claude-3.5-sonnet & 30.54  & 27.20  & 19.67 & 53.85 & 29.92 \\
    & o3-mini           & 30.96  & 71.55  & 13.81 & 37.43 & 28.57 \\
\cmidrule{1-7}
\multirow{5}{*}{ReAct} 
    & Llama-3.3         & 28.87  & 12.97  & 8.37  & 64.52 & 26.06 \\
    & DeepSeek-R1-70B   & 28.03  & 42.26  & 15.90 & 51.49 & 15 \\
    & GPT-4o            & 26.78  & 37.24  & 9.62  & 38.20 & 23.62 \\
    & Claude-3.5-sonnet & 35.56  & 12.55  & 11.30 & 66.67 & 35.71 \\
    & o3-mini           & 34.73  & 57.74  & 3.35  & 42.75 & 25.81 \\
\cmidrule{1-7}
\multicolumn{2}{l|}{Human experts} 
    & 66.53  & 66.53  & 0     & 71.07 & 57.50 \\
\bottomrule
\end{tabular}
\caption{Model performance comparison on the DiscoveryBench dataset. Samples with inexecutable code are excluded when calculating Accuracy (R=0).}
\label{tab:discoverybench}
\end{table*}

\begin{table*}[htbp]
\centering
\begin{tabular}{ll|rrr|rr}
\toprule
Prompt           & Model                & \begin{tabular}[c]{@{}r@{}}Overall\\ Accuracy\end{tabular} & \begin{tabular}[c]{@{}r@{}}Overall\\ Reproducibility\end{tabular} & \begin{tabular}[c]{@{}r@{}}Inexecutable\\ Code\end{tabular} & \begin{tabular}[c]{@{}r@{}}Accuracy\\ (R=1)\end{tabular} & \begin{tabular}[c]{@{}r@{}}Accuracy\\ (R=0)\end{tabular} \\
\midrule
\multicolumn{7}{c}{QRData} \\
\midrule
\multirow{5}{*}{CoT} 
    & Llama-3.3         & 52.67  & 30.28  & 8.14  & 68.91 & 51.65 \\
    & DeepSeek-R1-70B   & 53.69  & 55.47  & 12.47 & 63.76 & 57.14 \\
    & GPT-4o            & 48.85  & 48.85  & 10.69 & 59.38 & 49.06 \\
    & Claude-3.5-sonnet & 54.45  & 31.3   & 6.62  & 59.35 & 57.79 \\
    & o3-mini           & 58.52  & 75.06  & 2.54  & 58.31 & 65.91 \\
\cmidrule{1-7}
\multirow{5}{*}{RoT} 
    & Llama-3.3         & 54.2   & 37.15  & 6.36  & 71.23 & 49.1  \\
    & DeepSeek-R1-70B   & 51.65  & 55.47  & 16.03 & 65.60 & 53.57 \\
    & GPT-4o            & 50.89  & 51.4   & 7.38  & 64.85 & 42.59 \\
    & Claude-3.5-sonnet & 56.23  & 38.93  & 7.38  & 56.21 & 63.98 \\
    & o3-mini           & 62.09  & 81.68  & 1.53  & 63.55 & 60.61 \\
\cmidrule{1-7}
\multirow{5}{*}{RRefl.} 
    & Llama-3.3         & 52.93  & 49.87  & 6.36  & 66.84 & 44.77 \\
    & DeepSeek-R1-70B   & 55.73  & 66.41  & 10.43 & 61.69 & 63.74 \\
    & GPT-4o            & 51.15  & 67.43  & 5.09  & 58.87 & 41.67 \\
    & Claude-3.5-sonnet & 58.02  & 45.04  & 3.31  & 59.89 & 60.1  \\
    & o3-mini           & 59.54  & 90.33  & 1.27  & 59.72 & 66.67 \\
\cmidrule{1-7}
\multirow{5}{*}{ReAct} 
    & Llama-3.3         & 54.2   & 30.53  & 6.62  & 62.5  & 55.87 \\
    & DeepSeek-R1-70B   & 56.23  & 46.31  & 7.89  & 68.68 & 53.33 \\
    & GPT-4o            & 51.15  & 49.36  & 4.33  & 61.86 & 44.51 \\
    & Claude-3.5-sonnet & 55.22  & 27.23  & 2.8   & 53.27 & 58.18 \\
    & o3-mini           & 60.56  & 56.74  & 0.25  & 64.13 & 56.21 \\
\midrule
\multicolumn{7}{c}{StatQA} \\
\midrule
\multirow{5}{*}{CoT} 
    & Llama-3.3         & 71.25  & 63.75  & 8.5   & 81.18 & 70.27 \\
    & DeepSeek-R1-70B   & 69     & 66.75  & 16    & 84.64 & 72.46 \\
    & GPT-4o            & 79     & 78.5   & 4     & 83.76 & 75.71 \\
    & Claude-3.5-sonnet & 78.25  & 48     & 11    & 89.58 & 85.98 \\
    & o3-mini           & 77.5   & 81     & 3.5   & 80.56 & 79.03 \\
\cmidrule{1-7}
\multirow{5}{*}{RoT} 
    & Llama-3.3         & 72.5   & 67.75  & 7.5   & 80.07 & 73.74 \\
    & DeepSeek-R1-70B   & 63.5   & 64.25  & 18.25 & 83.66 & 55.71 \\
    & GPT-4o            & 79.25  & 81.75  & 3.75  & 85.02 & 67.24 \\
    & Claude-3.5-sonnet & 80.25  & 53.5   & 8     & 90.65 & 82.47 \\
    & o3-mini           & 78.5   & 84.25  & 3     & 81.01 & 80.39 \\
\cmidrule{1-7}
\multirow{5}{*}{RRefl.} 
    & Llama-3.3         & 72.75  & 76.5   & 7.25  & 81.7  & 63.08 \\
    & DeepSeek-R1-70B   & 74.25  & 77.75  & 9     & 83.92 & 67.92 \\
    & GPT-4o            & 82     & 90.25  & 2     & 83.66 & 83.87 \\
    & Claude-3.5-sonnet & 79     & 51.5   & 9.25  & 88.35 & 85.35 \\
    & o3-mini           & 79     & 94.25  & 1.5   & 80.37 & 76.47 \\
\cmidrule{1-7}
\multirow{5}{*}{ReAct} 
    & Llama-3.3         & 73.25  & 63.25  & 3.5   & 79.45 & 69.17 \\
    & DeepSeek-R1-70B   & 72.5   & 68.25  & 5.5   & 80.95 & 65.71 \\
    & GPT-4o            & 76.5   & 80.25  & 2.25  & 81.93 & 61.43 \\
    & Claude-3.5-sonnet & 84.75  & 47.75  & 1.25  & 90.05 & 81.86 \\
    & o3-mini           & 75.25  & 83.5   & 0.75  & 77.25 & 68.25 \\
\bottomrule
\end{tabular}
\caption{Model performance comparison on the QRData and StatQA datasets. Samples with inexecutable code are excluded when calculating Accuracy (R=0).}
\label{tab:qrdata-statqa}
\end{table*}

\begin{table*}[htbp]
  \centering
  \begin{tabular}{lcc|cc|cc}
    \toprule
                      & \multicolumn{2}{c}{All Datasets} & \multicolumn{2}{c}{StatQA} & \multicolumn{2}{c}{All Datasets} \\
                      & \multicolumn{2}{c}{(Prompt, LLM, Dataset)} & \multicolumn{2}{c}{(Prompt, LLM, Category)} & \multicolumn{2}{c}{(Prompt, LLM, Type)} \\
    \cmidrule(lr){2-3} \cmidrule(lr){4-5} \cmidrule(lr){6-7}
                      & Accuracy & Repr.   & Accuracy & Repr.   & Accuracy & Repr.   \\
    \cmidrule(lr){1-3} \cmidrule(lr){4-5} \cmidrule(lr){6-7}
    RoT               & 0.75     & 3.79    & -0.35    & 2.70    & 0.16     & 3.20    \\
    RReflexion         & 1.47     & 12.23\textsuperscript{***}   & 1.05     & 10.45\textsuperscript{***}   & 0.42     & 7.99\textsuperscript{**}    \\
    ReAct             & 2.16     & -3.25   & -0.80    & 1.00    & 0.83     & -3.29   \\
    \cmidrule(lr){1-3} \cmidrule(lr){4-5} \cmidrule(lr){6-7}
    DeepSeek-R1-70B   & -2.10    & 13.98\textsuperscript{***}   & 1.31     & 1.44    & 1.55     & 13.08\textsuperscript{***}   \\
    GPT-4o            & 0.42     & 18.57\textsuperscript{***}   & 5.63\textsuperscript{***}     & 14.88\textsuperscript{***}   & 0.91     & 17.04\textsuperscript{***}   \\
    Claude-3.5-sonnet & 4.09\textsuperscript{**}     & -6.63\textsuperscript{**}   & 10.38\textsuperscript{***}    & -17.63\textsuperscript{***}  & 5.71\textsuperscript{***}     & -8.72\textsuperscript{**}   \\
    o3-mini           & 5.12\textsuperscript{***}     & 31.64\textsuperscript{***}   & 4.56\textsuperscript{**}     & 17.94\textsuperscript{***}   & 4.91\textsuperscript{***}     & 29.27\textsuperscript{***}   \\
    \cmidrule(lr){1-3} \cmidrule(lr){4-5} \cmidrule(lr){6-7}
    QRData            & 26.18\textsuperscript{***}    & 12.10\textsuperscript{***}   &          &         &          &         \\
    StatQA            & 47.19\textsuperscript{***}    & 31.49\textsuperscript{***}   &          &         &          &         \\
    \cmidrule(lr){1-3} \cmidrule(lr){4-5} \cmidrule(lr){6-7}
    CTT               &          &         & 9.94\textsuperscript{***}     & 1.69    &          &         \\
    DS                &          &         & 27.38\textsuperscript{***}    & 16.50\textsuperscript{***}   &          &         \\
    DCT               &          &         & 28.69\textsuperscript{***}    & -1.38   &          &         \\
    VT                &          &         & 29.00\textsuperscript{***}    & -2.38   &          &         \\
    \cmidrule(lr){1-3} \cmidrule(lr){4-5} \cmidrule(lr){6-7}
    Categorical       &          &         &          &         & 1.63\textsuperscript{*}     & -4.27\textsuperscript{*}   \\
    Textual           &          &         &          &         & 2.06\textsuperscript{**}     & -15.09\textsuperscript{***}  \\
    \bottomrule
  \end{tabular}
  \caption{Coefficients of linear regression models across three settings. Each column represents a response variable, and each row represents a dependent variable. Empty cells indicate that the variable is not present in the corresponding model. *: $0.01\leq p<0.05$, **: $0.001 \leq p< 0.01$, ***: p<0.001. CoT is the baseline of the prompts. Llama-3.3 is the baseline of LLMs. DiscoveryBench is the baseline of datasets. CA is the baseline for statistical question categories. Numerical is the baseline for QA types.} 
  \label{tab:coef}
\end{table*}

\begin{table*}[htbp]
\centering
\begin{tabular}{|p{14cm}|}
\hline
\textbf{Question:} In which century did the Depots peak? \\
\textbf{LLM (Error Category - Sufficiency):} \\
1. Load the time series data from both files \\ 
2. Focus on the Depot-related columns \\ 
3. Find the century with the highest Depot value \\ 
4. Convert the time period to century format for better understanding \\
\textbf{Human:} \\
1. Import pandas as pd and matplotlib.pyplot as plt \\ 
2. Load the CSV file \texttt{time\_series\_data.csv} into a Pandas DataFrame \\ 
3. Select the relevant columns (CE, Depot\_inter) and drop rows with missing values \\ 
4. Identify the year of peak depot importance by finding the row with the maximum value in the column \texttt{Depot\_inter} \\ 
5. Extract the corresponding year (CE) for the peak depot importance \\ 
6. Print the year when depots peaked \\
\hline

\end{tabular}
\caption{\label{LLMvsHuman} An example showing the difference between LLM-generated and human-written workflow.}
\end{table*}

\begin{table*}[htbp]
\centering
\begin{tabular}{c l l c c}
\hline
Category & Model & Prompt & \begin{tabular}[c]{@{}c@{}}Overall\\ Accuracy\end{tabular} & \begin{tabular}[c]{@{}c@{}}Overall\\ Reproducibility\end{tabular} \\
\hline
\multirow{20}{*}{\rotatebox{90}{Correlation Analysis}} 
 & Llama-3.3         & CoT         & 48.75 & 57.50 \\
 & Llama-3.3         & RoT         & 55.00 & 61.25 \\
 & Llama-3.3         & RRefl.   & 51.25 & 72.50 \\
 & Llama-3.3         & ReAct       & 42.50 & 67.50 \\
 & DeepSeek-R1-70B   & CoT         & 60.00 & 60.00 \\
 & DeepSeek-R1-70B   & RoT         & 47.50 & 57.50 \\
 & DeepSeek-R1-70B   & RRefl.   & 63.75 & 75.00 \\
 & DeepSeek-R1-70B   & ReAct       & 55.00 & 63.75 \\
 & GPT-4o            & CoT         & 63.75 & 77.50 \\
 & GPT-4o            & RoT         & 61.25 & 78.75 \\
 & GPT-4o            & RRefl.   & 67.50 & 90.00 \\
 & GPT-4o            & ReAct       & 55.00 & 80.00 \\
 & Claude-3.5-sonnet & CoT         & 65.00 & 36.25 \\
 & Claude-3.5-sonnet & RoT         & 68.75 & 51.25 \\
 & Claude-3.5-sonnet & RRefl.   & 62.50 & 45.00 \\
 & Claude-3.5-sonnet & ReAct       & 70.00 & 46.25 \\
 & o3-mini           & CoT         & 62.50 & 80.00 \\
 & o3-mini           & RoT         & 67.50 & 80.00 \\
 & o3-mini           & RRefl.   & 61.25 & 93.75 \\
 & o3-mini           & ReAct       & 53.75 & 91.25 \\
\hline
\multirow{20}{*}{\rotatebox{90}{Contingency Table Test}}
 & Llama-3.3         & CoT         & 61.25 & 57.50 \\
 & Llama-3.3         & RoT         & 62.50 & 63.75 \\
 & Llama-3.3         & RRefl.   & 58.75 & 67.50 \\
 & Llama-3.3         & ReAct       & 63.75 & 61.25 \\
 & DeepSeek-R1-70B   & CoT         & 70.00 & 65.00 \\
 & DeepSeek-R1-70B   & RoT         & 66.25 & 61.25 \\
 & DeepSeek-R1-70B   & RRefl.   & 71.25 & 71.25 \\
 & DeepSeek-R1-70B   & ReAct       & 70.00 & 62.50 \\
 & GPT-4o            & CoT         & 65.00 & 76.25 \\
 & GPT-4o            & RoT         & 66.25 & 78.75 \\
 & GPT-4o            & RRefl.   & 70.00 & 87.50 \\
 & GPT-4o            & ReAct       & 62.50 & 75.00 \\
 & Claude-3.5-sonnet & CoT         & 81.25 & 58.75 \\
 & Claude-3.5-sonnet & RoT         & 82.50 & 60.00 \\
 & Claude-3.5-sonnet & RRefl.   & 83.75 & 61.25 \\
 & Claude-3.5-sonnet & ReAct       & 80.00 & 55.00 \\
 & o3-mini           & CoT         & 67.50 & 80.00 \\
 & o3-mini           & RoT         & 66.25 & 85.00 \\
 & o3-mini           & RRefl.   & 67.50 & 90.00 \\
 & o3-mini           & ReAct       & 65.00 & 81.25 \\
\hline
\end{tabular}
\caption{Results for Correlation Analysis and Contingency Table Test in StatQA}
\label{CACTT}
\end{table*}

\begin{table*}[htbp]
\centering
\begin{tabular}{c l l c c}
\hline
Category & Model & Prompt & \begin{tabular}[c]{@{}c@{}}Overall\\ Accuracy\end{tabular} & \begin{tabular}[c]{@{}c@{}}Overall\\ Reproducibility\end{tabular} \\
\hline
\multirow{20}{*}{\rotatebox{90}{Descriptive Statistics}}
 & Llama-3.3         & CoT         & 82.50 & 85.00 \\
 & Llama-3.3         & RoT         & 82.50 & 82.50 \\
 & Llama-3.3         & RRefl.   & 85.00 & 90.00 \\
 & Llama-3.3         & ReAct       & 83.75 & 72.50 \\
 & DeepSeek-R1-70B   & CoT         & 80.00 & 82.50 \\
 & DeepSeek-R1-70B   & RoT         & 80.00 & 86.25 \\
 & DeepSeek-R1-70B   & RRefl.   & 83.75 & 91.25 \\
 & DeepSeek-R1-70B   & ReAct       & 86.25 & 82.50 \\
 & GPT-4o            & CoT         & 90.00 & 81.25 \\
 & GPT-4o            & RoT         & 88.75 & 85.00 \\
 & GPT-4o            & RRefl.   & 88.75 & 92.50 \\
 & GPT-4o            & ReAct       & 86.25 & 93.75 \\
 & Claude-3.5-sonnet & CoT         & 92.50 & 73.75 \\
 & Claude-3.5-sonnet & RoT         & 91.25 & 78.75 \\
 & Claude-3.5-sonnet & RRefl.   & 86.25 & 76.25 \\
 & Claude-3.5-sonnet & ReAct       & 92.50 & 71.25 \\
 & o3-mini           & CoT         & 87.50 & 90.00 \\
 & o3-mini           & RoT         & 87.50 & 91.25 \\
 & o3-mini           & RRefl.   & 87.50 & 98.75 \\
 & o3-mini           & ReAct       & 87.50 & 90.00 \\
\hline
\multirow{20}{*}{\rotatebox{90}{Distribution Compliance Test}}
 & Llama-3.3         & CoT         & 87.50 & 61.25 \\
 & Llama-3.3         & RoT         & 86.25 & 72.50 \\
 & Llama-3.3         & RRefl.   & 87.50 & 77.50 \\
 & Llama-3.3         & ReAct       & 90.00 & 52.50 \\
 & DeepSeek-R1-70B   & CoT         & 83.75 & 53.75 \\
 & DeepSeek-R1-70B   & RoT         & 82.50 & 51.25 \\
 & DeepSeek-R1-70B   & RRefl.   & 87.50 & 75.00 \\
 & DeepSeek-R1-70B   & ReAct       & 77.50 & 62.50 \\
 & GPT-4o            & CoT         & 87.50 & 82.50 \\
 & GPT-4o            & RoT         & 88.75 & 82.50 \\
 & GPT-4o            & RRefl.   & 90.00 & 91.25 \\
 & GPT-4o            & ReAct       & 90.00 & 88.75 \\
 & Claude-3.5-sonnet & CoT         & 88.75 & 33.75 \\
 & Claude-3.5-sonnet & RoT         & 87.50 & 37.50 \\
 & Claude-3.5-sonnet & RRefl.   & 90.00 & 36.25 \\
 & Claude-3.5-sonnet & ReAct       & 91.25 & 35.00 \\
 & o3-mini           & CoT         & 91.25 & 77.50 \\
 & o3-mini           & RoT         & 88.75 & 88.75 \\
 & o3-mini           & RRefl.   & 93.75 & 95.00 \\
 & o3-mini           & ReAct       & 86.25 & 82.50 \\
\hline
\end{tabular}
\caption{Results for Descriptive Statistics and Distribution Compliance Test in StatQA}
\label{DSDCT}
\end{table*}

\begin{table*}[htbp]
\centering
\begin{tabular}{c l l c c}
\hline
Category & Model & Prompt & \begin{tabular}[c]{@{}c@{}}Overall\\ Accuracy\end{tabular} & \begin{tabular}[c]{@{}c@{}}Overall\\ Reproducibility\end{tabular} \\
\hline
\multirow{20}{*}{\rotatebox{90}{Variance Test}}
 & Llama-3.3         & CoT         & 85.00 & 57.50 \\
 & Llama-3.3         & RoT         & 82.50 & 58.75 \\
 & Llama-3.3         & RRefl.   & 86.25 & 75.00 \\
 & Llama-3.3         & ReAct       & 92.50 & 62.50 \\
 & DeepSeek-R1-70B   & CoT         & 83.75 & 72.50 \\
 & DeepSeek-R1-70B   & RoT         & 83.75 & 65.00 \\
 & DeepSeek-R1-70B   & RRefl.   & 83.75 & 76.25 \\
 & DeepSeek-R1-70B   & ReAct       & 85.00 & 70.00 \\
 & GPT-4o            & CoT         & 90.00 & 75.00 \\
 & GPT-4o            & RoT         & 92.50 & 83.75 \\
 & GPT-4o            & RRefl.   & 93.75 & 90.00 \\
 & GPT-4o            & ReAct       & 90.00 & 63.75 \\
 & Claude-3.5-sonnet & CoT         & 92.50 & 37.50 \\
 & Claude-3.5-sonnet & RoT         & 92.50 & 40.00 \\
 & Claude-3.5-sonnet & RRefl.   & 91.25 & 38.75 \\
 & Claude-3.5-sonnet & ReAct       & 92.50 & 31.25 \\
 & o3-mini           & CoT         & 86.25 & 77.50 \\
 & o3-mini           & RoT         & 86.25 & 76.25 \\
 & o3-mini           & RRefl.   & 87.50 & 93.75 \\
 & o3-mini           & ReAct       & 85.00 & 72.50 \\
\hline
\end{tabular}
\caption{Results for Variance Test in StatQA}
\label{VT}
\end{table*}

\begin{figure*}[htbp]
  \includegraphics[width=0.95\linewidth]{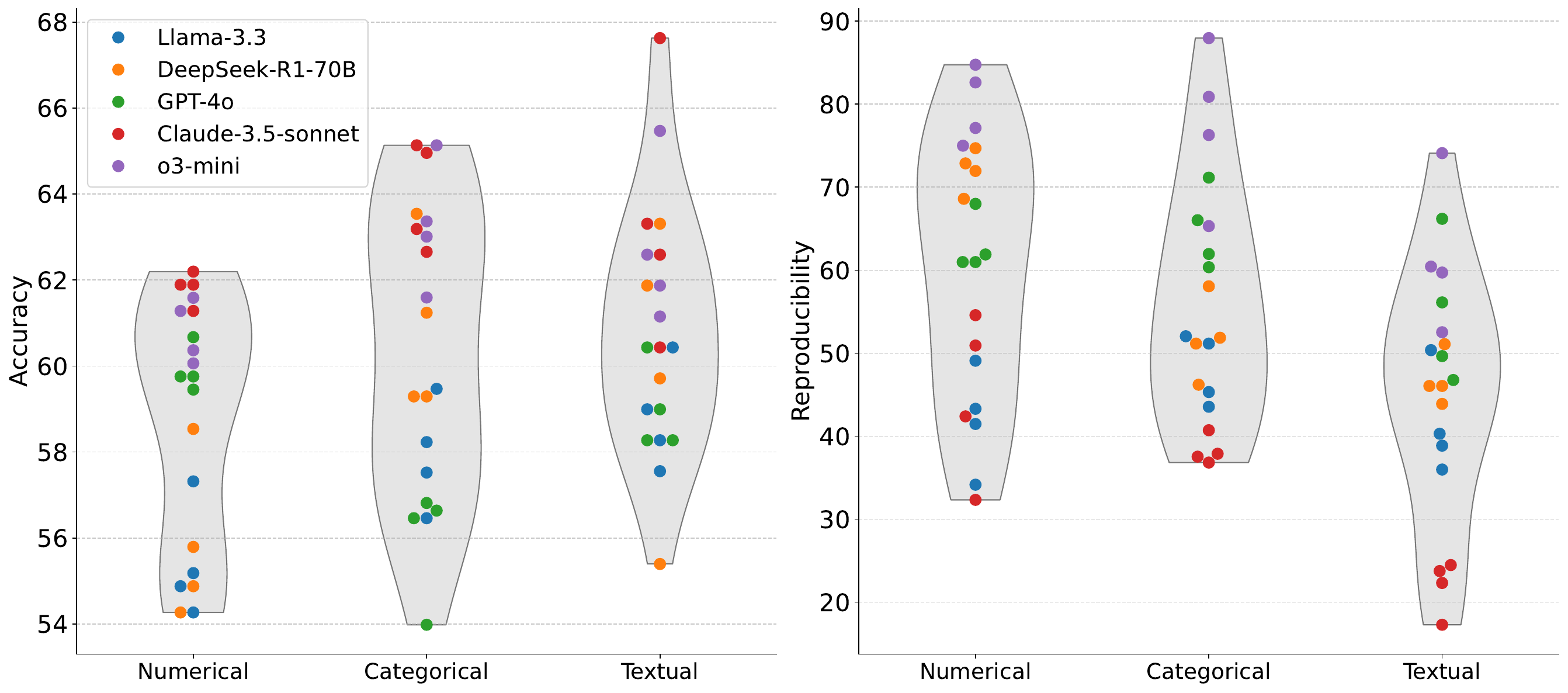} 
  \caption {Accuracy and reproducibility of LLMs for different QA types across all three benchmark datasets. 
  }
  \label{Ctgr}
\end{figure*}

\begin{figure*}[t]
\centering
\includegraphics[width=\linewidth]{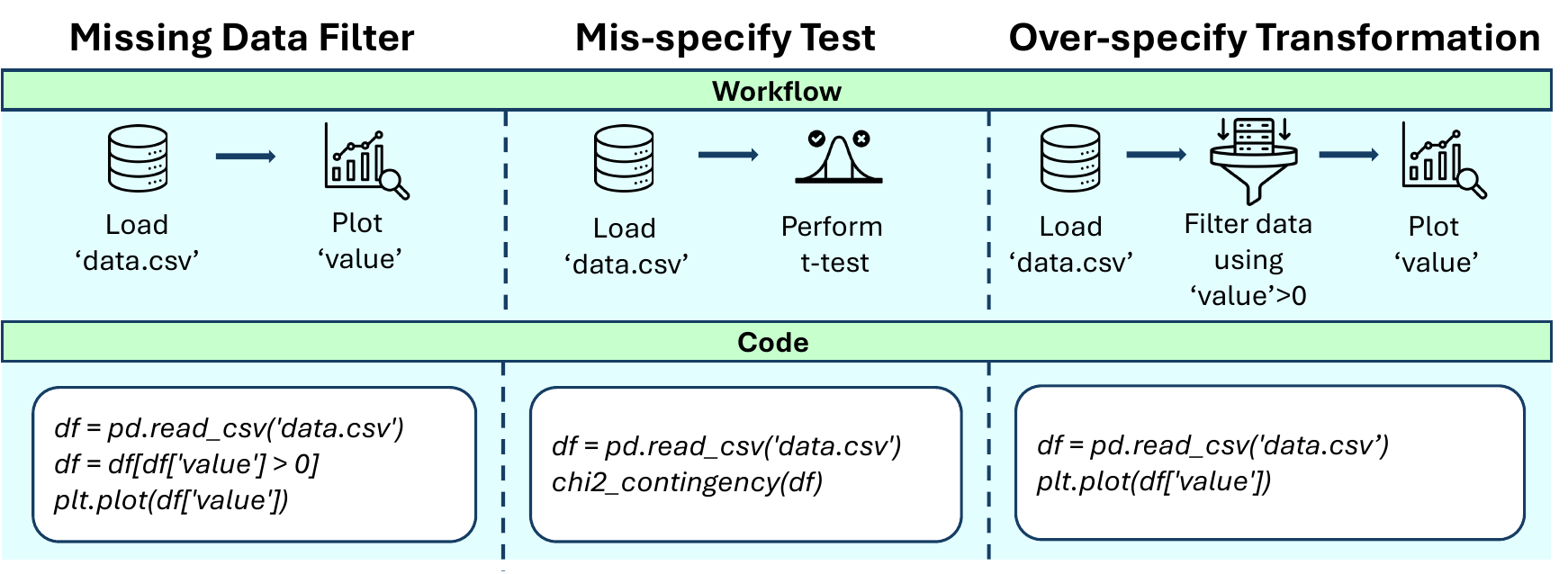}
  \caption{Examples of three types of irreproducibility.}
\label{error_example}
\end{figure*}

\begin{figure*}[t]
\centering
\includegraphics[width=\linewidth]{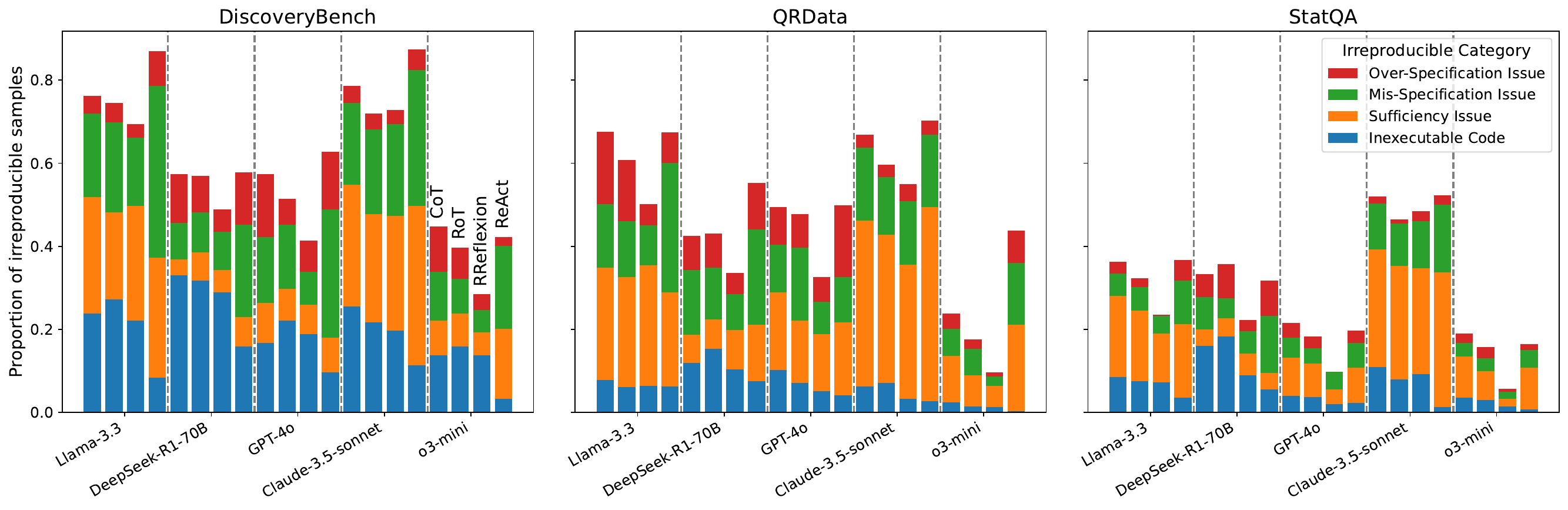}
  \caption{Proportion of irreproducible samples by category for five LLMs in three benchmark datasets.}
\label{Error_Analysis}
\end{figure*}

\begin{figure*}[htbp]
  \includegraphics[width=0.95\linewidth]{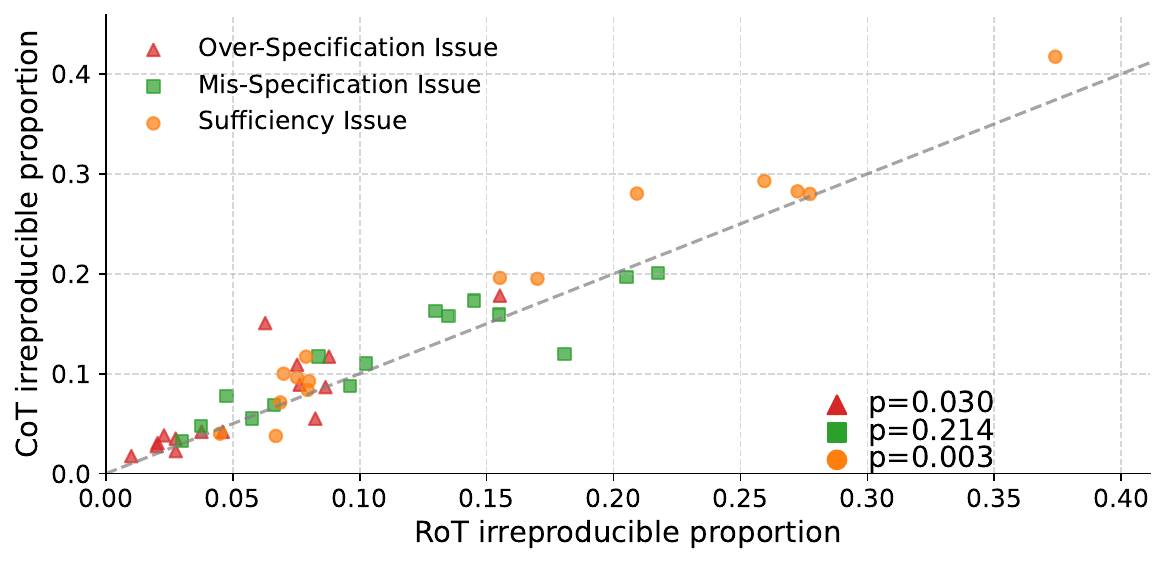} 
  \caption {Comparison between CoT and RoT on different types of irreproducibility. P values are obtained from two sample one-sided (RoT $<$ CoT) t-tests. 
  }
  \label{error_analysis_CoT_vs_RoT}
\end{figure*}

\begin{figure*}[htbp]
  \includegraphics[width=0.95\linewidth]{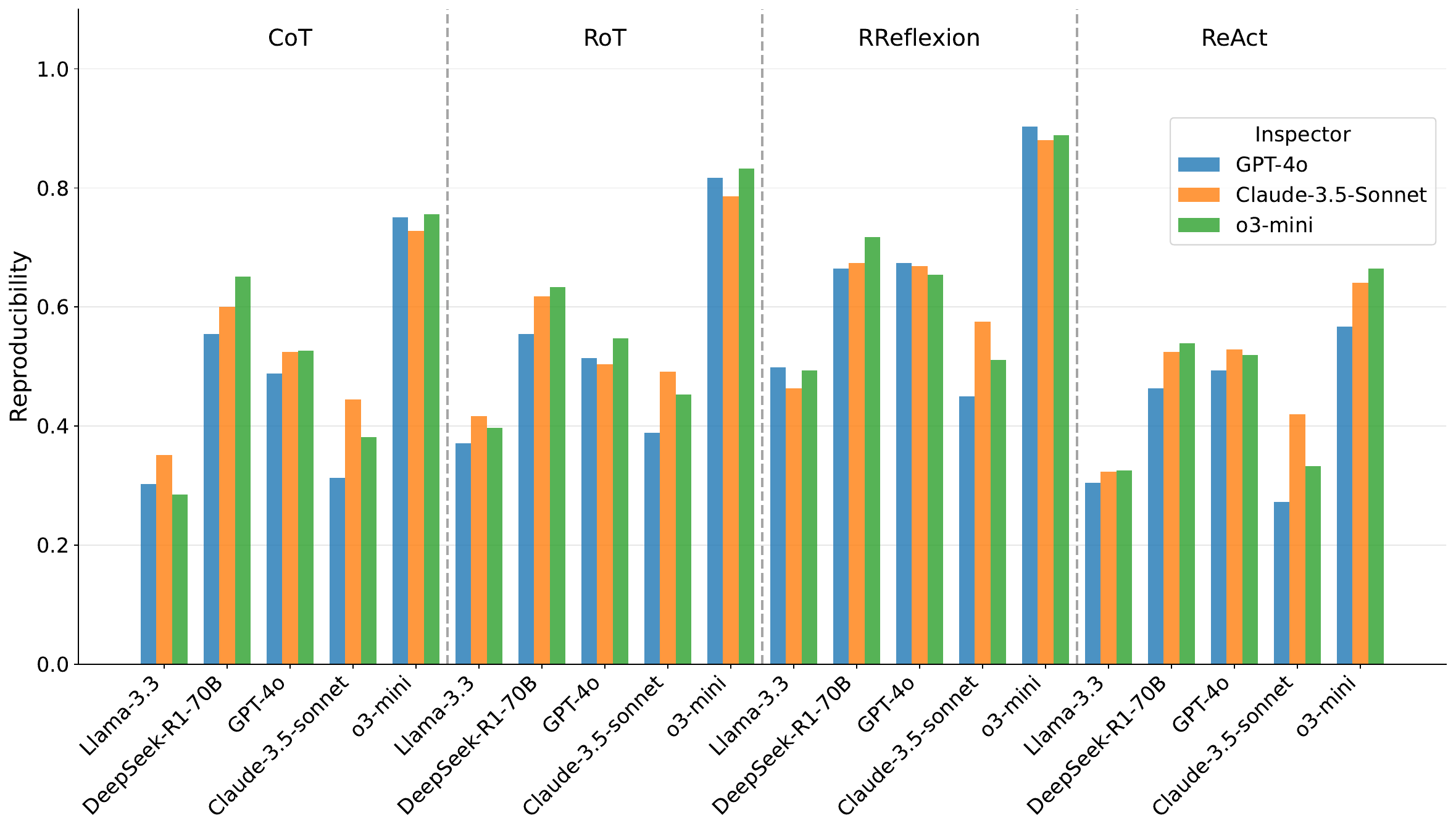} 
  \caption {Reproducibility scores on QRData with GPT-4o, Claude-3.5-sonnet, and o3-mini as the inspectors.}
  \label{inspector}
\end{figure*}

\begin{table*}[htbp]
\centering
\begin{tabular}{c l l c c}
\hline
QA Type & Model & Prompt & \begin{tabular}[c]{@{}c@{}}Overall\\ Accuracy\end{tabular} & \begin{tabular}[c]{@{}c@{}}Overall\\ Reproducibility\end{tabular} \\
\hline
\multirow{20}{*}{\rotatebox{90}{Numerical}}
 & Llama-3.3         & CoT         & 54.27 & 41.46 \\
 & Llama-3.3         & RoT         & 54.88 & 43.29 \\
 & Llama-3.3         & RRefl.   & 55.18 & 49.09 \\
 & Llama-3.3         & ReAct       & 57.32 & 34.15 \\
 & DeepSeek-R1-70B   & CoT         & 54.27 & 72.87 \\
 & DeepSeek-R1-70B   & RoT         & 54.88 & 71.95 \\
 & DeepSeek-R1-70B   & RRefl.   & 55.79 & 74.70 \\
 & DeepSeek-R1-70B   & ReAct       & 58.54 & 68.60 \\
 & GPT-4o            & CoT         & 59.76 & 60.98 \\
 & GPT-4o            & RoT         & 59.76 & 60.98 \\
 & GPT-4o            & RRefl.   & 59.45 & 67.99 \\
 & GPT-4o            & ReAct       & 60.67 & 61.89 \\
 & Claude-3.5-sonnet & CoT         & 62.20 & 42.38 \\
 & Claude-3.5-sonnet & RoT         & 61.89 & 54.57 \\
 & Claude-3.5-sonnet & RRefl.   & 61.28 & 50.91 \\
 & Claude-3.5-sonnet & ReAct       & 61.89 & 32.32 \\
 & o3-mini           & CoT         & 60.06 & 75.00 \\
 & o3-mini           & RoT         & 61.59 & 82.62 \\
 & o3-mini           & RRefl.   & 60.37 & 84.76 \\
 & o3-mini           & ReAct       & 61.28 & 77.13 \\
\hline
\multirow{20}{*}{\rotatebox{90}{Categorical}}
 & Llama-3.3         & CoT         & 58.23 & 45.31 \\
 & Llama-3.3         & RoT         & 59.47 & 52.04 \\
 & Llama-3.3         & RRefl.   & 56.46 & 51.15 \\
 & Llama-3.3         & ReAct       & 57.52 & 43.54 \\
 & DeepSeek-R1-70B   & CoT         & 61.24 & 51.86 \\
 & DeepSeek-R1-70B   & RoT         & 59.29 & 51.15 \\
 & DeepSeek-R1-70B   & RRefl.   & 63.54 & 58.05 \\
 & DeepSeek-R1-70B   & ReAct       & 59.29 & 46.19 \\
 & GPT-4o            & CoT         & 56.81 & 61.95 \\
 & GPT-4o            & RoT         & 56.46 & 66.02 \\
 & GPT-4o            & RRefl.   & 56.64 & 71.15 \\
 & GPT-4o            & ReAct       & 53.98 & 60.35 \\
 & Claude-3.5-sonnet & CoT         & 62.65 & 36.81 \\
 & Claude-3.5-sonnet & RoT         & 65.13 & 40.71 \\
 & Claude-3.5-sonnet & RRefl.   & 63.19 & 37.88 \\
 & Claude-3.5-sonnet & ReAct       & 64.96 & 37.52 \\
 & o3-mini           & CoT         & 63.01 & 76.28 \\
 & o3-mini           & RoT         & 65.13 & 80.88 \\
 & o3-mini           & RRefl.   & 63.36 & 87.96 \\
 & o3-mini           & ReAct       & 61.59 & 65.31 \\
\hline
\end{tabular}
\caption{Results for Categorical and Numerical QA types across all three datasets.}
\label{CatNum}
\end{table*}

\begin{table*}[htbp]
\centering
\begin{tabular}{c l l c c}
\hline
QA Type & Model & Prompt & \begin{tabular}[c]{@{}c@{}}Overall\\ Accuracy\end{tabular} & \begin{tabular}[c]{@{}c@{}}Overall\\ Reproducibility\end{tabular} \\
\hline
\multirow{20}{*}{\rotatebox{90}{Textual}}
 & Llama-3.3         & CoT         & 58.27 & 35.97 \\
 & Llama-3.3         & RoT         & 58.99 & 38.85 \\
 & Llama-3.3         & RRefl.   & 57.55 & 50.36 \\
 & Llama-3.3         & ReAct       & 60.43 & 40.29 \\
 & DeepSeek-R1-70B   & CoT         & 59.71 & 46.04 \\
 & DeepSeek-R1-70B   & RoT         & 55.40 & 43.88 \\
 & DeepSeek-R1-70B   & RRefl.   & 63.31 & 51.08 \\
 & DeepSeek-R1-70B   & ReAct       & 61.87 & 46.04 \\
 & GPT-4o            & CoT         & 58.27 & 49.64 \\
 & GPT-4o            & RoT         & 58.27 & 56.12 \\
 & GPT-4o            & RRefl.   & 60.43 & 66.19 \\
 & GPT-4o            & ReAct       & 58.99 & 46.76 \\
 & Claude-3.5-sonnet & CoT         & 63.31 & 22.30 \\
 & Claude-3.5-sonnet & RoT         & 60.43 & 24.46 \\
 & Claude-3.5-sonnet & RRefl.   & 62.59 & 23.74 \\
 & Claude-3.5-sonnet & ReAct       & 67.63 & 17.27 \\
 & o3-mini           & CoT         & 62.59 & 60.43 \\
 & o3-mini           & RoT         & 65.47 & 59.71 \\
 & o3-mini           & RRefl.   & 61.87 & 74.10 \\
 & o3-mini           & ReAct       & 61.15 & 52.52 \\
\hline
\end{tabular}
\caption{Results for Textual QA types across all three datasets.}
\label{TXT}
\end{table*}

\begin{table*}
  \centering
  \begin{tabular}{p{0.95\textwidth}}
    \hline
    You are a statistician trying to answer a question based on one or more datasets.\\ [1ex]

    You have access to the following tools: \\
    python\_repl\_ast: A Python shell. Use this to execute python commands. Input should be a valid python command. When using this tool, sometimes output is abbreviated - make sure it does not look abbreviated before using it in your answer. \\

    In your output, please strictly follow the format outlined below, maintaining the specified order and structure.\\

    Question: the input question you must answer\\
    Workflow: a plan to tackle the problem\\
    Action: the action to take, should be one of [python\_repl\_ast]\\
    Action Input: the input to the action in the format of (\verb|```python\s.*?```|). Use print to show the results.\\
    Observation: the result of performing the action with the action input (please do not generate)\\
    Final Answer: an answer to the original question\\ [1ex]

    NOTE: You will need to generate the complete action input in one code snippet to solve the query in one attempt. \\
    If no observation is provided, you need to generate the workflow, action, and action input. You don't need to provide the final answer. \\
    If an observation is provided, you should generate the answer starting with "Final Answer:" \{agent\_instruction\}\\ [1ex]

    Begin!\\ [1ex]

    You need to load all datasets in Python using the specified paths: \\
    \{file\_paths\}\\ [1ex]

    Dataset descriptions: \\
    \{descriptions\}\\ [1ex]

    Question: \\
    \{question\}\\
    \{conversation\}\{reflexion\}\\
    \hline
  \end{tabular}
  \caption{\label{CoTPrompt}
    Prompt templates for Chain-of-Thought (CoT), Reproducibility-of-Thought (RoT), and Reproducibility-Reflexion (RReflexion). For the CoT prompt, $\text{\{agent\_instruction\}}$ is replaced with "\textbackslash nLet's think step by step." The RoT prompt adds the instruction: "Make sure a person can replicate the action input by only looking at the workflow, and the action input reflects every step of the workflow." For the RReflexion prompt, $\text{\{reflexion\}}$ is replaced with "\textbackslash nThe above workflow and action input are not aligned for reproducibility. Please rewrite the workflow, action, and action input. Generate the most likely executable code (action input) and ensure reproducibility of the code through the workflow."
  }
\end{table*}

\begin{table*}
  \centering
  \begin{tabular}{p{0.95\textwidth}}
    \hline
    You are a statistician trying to answer a question based on one or more datasets. \\ [1ex]

You have access to the following tools: \\
python\_repl\_ast: A Python shell. Use this to execute python commands. Input should be a valid python command. When using this tool, sometimes output is abbreviated - make sure it does not look abbreviated before using it in your answer.\\[1ex]

Below is the structure of the agent-environment interaction. Your task is to generate only the agent's responses. \\
Question: the input question you must answer \\
<Agent> \\
Workflow: a plan to tackle the problem \\
Action: the action to take, should be one of [python\_repl\_ast] \\
Action Input: the input to the action in the format of (\verb|```python\s.*?```|). Use print to show the results. \\
<Environment> \\
Observation: the result of performing the action with the action input (please do not generate) \\
<Agent> \\
Workflow: a plan to tackle the problem \\
Action: the action to take, should be one of [\{tool\_names\}] \\
Action Input: the input to the action in the format of (\verb|```python\s.*?```|). Use print to show the results. \\
<Environment> \\
Observation: the result of performing the action with the action input (please do not generate) \\
...  \\
<Agent> \\
Final Answer: an answer to the original question \\
Task done! \\[1ex]

NOTE: You will need to generate the complete action input in one code snippet. We will execute the code for you and provide the observation to you. \\
If no observation is provided or the observation is insufficient to answer the question, you need to (re)generate the complete workflow, action, and action input. Do not generate final answer and 'Task done'. \\
If the observation is sufficient to answer the question, generate the final answer starting with 'Final Answer:' and indicate completion by ending with 'Task done!'. Do not generate workflow any more. \\
Let's think step by step.\\[1ex]

Example \\
Question:  \\
What is the linear coefficient that describes the positive relationship between the rate of maximum body length evolution and spatial variation in speciation rates, where the former emerges as the most influential factor? \\
<Agent> \\
Workflow: First, I need to load the dataset 'body-size-evolution-in-south-american-freshwater-fishes.csv' into a Pandas DataFrame, and print the first few rows of the data. \\
Action: python\_repl\_ast \\
Action Input: \verb|```python|\begin{verbatim}
import pandas as pd
df = pd.read_csv('body-size-evolution-in-south-american-freshwater-fishes.csv')
df.head()
\end{verbatim}
\verb|```|\\
    \hline
  \end{tabular}
  \caption{\label{ReActPrompt}
    Prompt template for ReAct.
  }
\end{table*}

\begin{table*}
  \centering
  \begin{tabular}{p{0.95\textwidth}}
    \hline
    <Environment> \\
    Observation: \\
     HYBAS\_ID       long       lat  BAMM\_speciation  BAMM\_extinction  BAMM\_NetDiv  ...   aet    Elevation  sgr  soil\_div     area  diversity \\
0  6050000010 -76.477422  7.742693         0.137392         0.026807     0.110585  ...  1387   330.150088  166  0.482402  72363.7         68\\
1  6050000740 -74.628725  9.803586         0.117235         0.025796     0.091438  ...  1082    69.475294   23  0.457436  17944.3         35\\
2  6050068100 -75.295995  8.448815         0.119381         0.023826     0.095555  ...  1312   143.032178   74  0.378793  17105.5         44\\
3  6050068110 -74.608408  8.922863         0.132477         0.027777     0.104700  ...  1445    14.724138    3  0.468328    610.1         48\\
4  6050070260 -75.591588  5.770093         0.120127         0.022940     0.097187  ...  1371  1378.729945  421  0.158870  61901.9         81\\
\text{[5 rows x 21 columns]} \\
<Agent>\\
Workflow: 1. Import Libraries: Import `LinearRegression` from `sklearn` and `numpy`.\\
2. Prepare Data: Extract predictor 'BAMM\_speciation' and response 'BAMM\_NetDiv'.\\
3. Create and Fit Model: Instantiate `LinearRegression()` and fit the model using `model.fit(X, y)`.\\
4. Retrieve Coefficient: Get the slope using `model.coef\_[0]`.\\
5. Output Result: Print the coefficient with `print(coefficient)`.\\
Action: python\_repl\_ast\\
Action Input: \verb|```python|\begin{verbatim}
from sklearn.linear_model import LinearRegression
import numpy as np

# Prepare the data for regression
X = df[['BAMM_speciation']].values.reshape(-1, 1)  # Predictor
y = df['BAMM_NetDiv'].values  # Response

# Create a linear regression model
model = LinearRegression()
model.fit(X, y)

# Get the coefficient
coefficient = model.coef_[0]
print(coefficient)
\end{verbatim}
\verb|```|\\
<Environment>\\
Observation: 0.5175306498596297\\
<Agent>\\
Final Answer:\\
The linear coefficient that describes the positive relationship between the rate of maximum body length evolution ('BAMM\_speciation') and spatial variation in speciation rates ('BAMM\_NetDiv') is approximately 0.518.\\
Task done!\\[1ex]

    \hline
  \end{tabular}
  \caption{\label{ContinueReActPrompt}
    [Continue] Prompt template for ReAct.
  }
\end{table*}

\begin{table*}
  \centering
  \begin{tabular}{p{0.95\textwidth}}
    \hline
Begin!\\[1ex]

You need to load all datasets in python using the specified paths: \\
\{file\_paths\}\\[1ex]

Dataset descriptions:\\ 
{descriptions}\\[1ex]

Question: \\
\{question\}\\
<Agent>\\
\{conversation\}\\
    \hline
  \end{tabular}
  \caption{\label{Continue2ReActPrompt}
    [Continue] Prompt template for ReAct.
  }
\end{table*}

\begin{table*}
  \centering
  \begin{tabular}{p{0.95\textwidth}}
    \hline
    Evaluate the correctness (0 for incorrect, 1 for correct) of the predicted answer to the question: \{question\}\\
\{dataset\_specific\_prompt\}\\
Predicted answer: \{predicted\_answer\}\\[1ex]

Ground truth answer: \{true\_answer\}\\[1ex]

Please reply in this format: \\
"Thoughts: \\[1ex]

The accuracy score is:"\\
    \hline
  \end{tabular}
  \caption{\label{AccuracyPrompt}
    Prompt for evaluating the accuracy of an answer. The $\text{\{dataset\_specific\_prompt}\}$ is used to define dataset-specific requirements. For questions with a numerical conclusion, accuracy is considered achieved if the deviation between LLM’s response and the benchmark conclusion falls below a benchmark-specific error threshold: 3\% threshold for QRData following \citep{liu2024llms} and 1\% threshold for DiscoveryBench and StatQA. If the conclusions in StatQA tasks contradict each other, any conclusion is considered correct.
  }
\end{table*}

\begin{table*}
  \centering
  \begin{tabular}{p{0.95\textwidth}}
    \hline
    Question: \{question\}\\
These are the dataset paths: \{file\_paths\}\\[1ex]

Develop a Python script that precisely converts the provided narrative summary into executable code. Ensure that each component of the analysis process is correctly implemented, closely following the steps outlined in the summary. Maintain consistency by using the exact variable names specified in the narrative. Below is the code summary to translate:\\
\{workflow\}\\
    \hline
  \end{tabular}
  \caption{\label{WorkflowCodePrompt}
    Prompt for workflow-to-code conversion. 
  }
\end{table*}

\begin{table*}
  \centering
  \begin{tabular}{p{0.95\textwidth}}
    \hline
    You are a data scientist answering a question based on the code and code output.\\[1ex]

    Question: \{question\}\\[1ex]
    
    Code:\\
    \{code\}\\[1ex]
    
    Code output: \{code\_output\}\\[1ex]
    
    Please generate the answer based only on the code output in this format: \\
    "Thought:\\[1ex]
    
    Conclusion:"\\
    \hline
  \end{tabular}
  \caption{\label{ConclusionPrompt}
    Prompt for extracting conclusion from code execution output. 
  }
\end{table*}

\begin{table*}
  \centering
  \begin{tabular}{p{0.95\textwidth}}
    \hline
    Your task is to determine whether Code 1 and Code 2 arrive at the same conclusion regarding the question: \{question\}\\[1ex]

Code 1:\\
\{code\_1\}\\[1ex]

Code 1 execution output:\\
\{code\_1\_output\}\\[1ex]

Code 1 conclusion:\\
\{code\_1\_conclusion\}\\[1ex]

Code 2:\\
\{code\_2\}\\[1ex]

Code 2 execution output:\\
\{code\_2\_output\}\\[1ex]

Code 2 conclusion:\\
\{code\_2\_conclusion\}\\[1ex]

If the output of the two code snippets provide the same values for the same statistics and lead to the same conclusion to the question, please score 1. Otherwise, score 0. Note that if code 2 runs into error, please score 0.
Please reply in this format: \\
"Thoughts: \\[1ex]

The similarity score is:"\\
    \hline
  \end{tabular}
  \caption{\label{ReprPrompt}
    Prompt for assessing reproducibility. 
  }
\end{table*}

\begin{table*}
  \centering
  \begin{tabular}{p{0.95\textwidth}}
    \hline
    I used an LLM to write a workflow and code1. Then the workflow is converted to code2.\\
    I did reproducibility test between code1 and the workflow-converted code2 to see if they are functionally equivalent.\\
    Analyze the following reproducibility test result and categorize this sample.\\ [1ex]

    Original code1: \\
    \verb|```| \\
    \{code1\} \\
    \verb|```| \\ [1ex]

    Original workflow: \\
    \{workflow\} \\ [1ex]

    Workflow-converted code2: \\
    \verb|```| \\
    \{code2\} \\
    \verb|```| \\ [1ex]

    Not reproducible reason: \{reason\} \\ [1ex]

    Assign exactly one category from the following list: \\
    1. "Workflow Sufficiency Issues": This occurs when the workflow leaves out details of the original analysis or oversimplifies the methods in the original code. \\
    Example: The workflow neglects to include variable names or normalization steps that were crucial in the original data analysis. \\
    2. "Workflow Mis-Specification Issue": This arises when the workflow inaccurately describes some steps. \\
    Example: The workflow mistakenly specifies using a logistic regression model, while the original code actually implements a linear regression model. \\
    3. "Workflow Over-Specification Issue": This happens when the workflow introduces additional steps, constraints, or assumptions that are not part of the original code. \\
    Example: The workflow unnecessarily adds an outlier-removal procedure that wasn't performed in the original analysis. \\ [1ex]

    Analyze what issues exist in the workflow that prevented it from accurately capturing the functionality of code1. Focus on how the workflow itself may be incomplete, incorrect, or unnecessarily complex rather than just comparing the code implementations. \\ [1ex]

    Your output must include: \\
    1. The reason for assigning this category, focusing on workflow issues \\
    2. The single category for this sample (chosen from the three categories above) \\ [1ex]

    \{format\_instructions\} \\
    \hline
  \end{tabular}
  \caption{\label{WorkflowReproPrompt}
    Prompt template for categorizing workflow reproducibility issues.
  }
\end{table*}

\end{document}